\documentclass[11pt, a4paper, onecolumn, copyright, goog]{google}

\usepackage{multirow}
\usepackage[square, numbers]{natbib}
\uselogo{}

\title{EmbeddingGemma 2: Unified and Efficient Multimodal Representations}
\renewcommand{\today}{}

\author[*]{Henrique Schechter Vera}
\author[*]{Sahil Dua}
\author[ \hspace{-0.3em}]{Jovan Andonov}
\author[ \hspace{-0.3em}]{Sara Smoot}
\author[ \hspace{-0.3em}]{Sebastian Russo}
\author[ \hspace{-0.3em}]{Ivan Llanos}
\author[ \hspace{-0.3em}]{Max Matveev}
\author[ \hspace{-0.3em}]{Jae Hyeon Yoo}
\author[ \hspace{-0.3em}]{Kat Black}
\author[ \hspace{-0.3em}]{Shih-Cheng Huang}
\author[ \hspace{-0.3em}]{Josiah Aklilu}
\author[ \hspace{-0.3em}]{Maarten Grootendorst}
\author[ \hspace{-0.3em}]{Omar Sanseviero}
\author[ \hspace{-0.3em}]{Chandramouli Amarnath}
\author[ \hspace{-0.3em}]{Ian Ballantyne}
\author[ \hspace{-0.3em}]{Zhe Li}
\author[ \hspace{-0.3em}]{Daniel Salz}
\author[ \hspace{-0.3em}]{Ibrahim Alabdulmohsin}
\author[ \hspace{-0.3em}]{Ilya Philippov}
\author[ \hspace{-0.3em}]{Kristen Quan}
\author[ \hspace{-0.3em}]{Glenn Cameron}
\author[ \hspace{-0.3em}]{Edouard Yvinec}
\author[ \hspace{-0.3em}]{Luciano Martins}
\author[ \hspace{-0.3em}]{Doug Reid}
\author[ \hspace{-0.3em}]{Sindhu Raghuram Panyam}
\author[ \hspace{-0.3em}]{Weiyi Wang}
\author[ \hspace{-0.3em}]{Hengxuan Ying}
\author[ \hspace{-0.3em}]{Madhuri Shanbhogue}
\author[ \hspace{-0.3em}]{Tanmaya Dabral}
\author[ \hspace{-0.3em}]{Shanfeng Zhang}
\author[ \hspace{-0.3em}]{Juyeong Ji}
\author[ \hspace{-0.3em}]{Renjie Wu}
\author[ \hspace{-0.3em}]{Joe Zou}
\author[ \hspace{-0.3em}]{Xu Chen}
\author[ \hspace{-0.3em}]{Yunan Zhang}
\author[ \hspace{-0.3em}]{Yu-Hui Chen}
\author[ \hspace{-0.3em}]{Matthew Soulanille}
\author[ \hspace{-0.3em}]{Vinod Mamillapalli}
\author[ \hspace{-0.3em}]{Yi-Chun Kuo}
\author[ \hspace{-0.3em}]{Wai Hon Law}
\author[ \hspace{-0.3em}]{Matt Kreileder}
\author[ \hspace{-0.3em}]{Alice Zheng}
\author[ \hspace{-0.3em}]{Suril Shah}
\author[ \hspace{-0.3em}]{Mogan Shieh}
\author[ \hspace{-0.3em}]{Pooja Consul}
\author[ \hspace{-0.3em}]{Alberto Montes}
\author[ \hspace{-0.3em}]{Andreas Hess}
\author[ \hspace{-0.3em}]{Andres Sevilla}
\author[ \hspace{-0.3em}]{Babak Samari}
\author[ \hspace{-0.3em}]{Brendan McCloskey}
\author[ \hspace{-0.3em}]{Brian Potetz}
\author[ \hspace{-0.3em}]{Daniel Cer}
\author[ \hspace{-0.3em}]{Dilip Kumar Margam}
\author[ \hspace{-0.3em}]{Feodor Beloborodov}
\author[ \hspace{-0.3em}]{Georgios Kathareios}
\author[ \hspace{-0.3em}]{Gustavo Hernández Ábrego}
\author[ \hspace{-0.3em}]{Hesen Zhang}
\author[ \hspace{-0.3em}]{Ji Ma}
\author[ \hspace{-0.3em}]{Jianan Lu}
\author[ \hspace{-0.3em}]{Jiaxing Wu}
\author[ \hspace{-0.3em}]{Keegan Mosley}
\author[ \hspace{-0.3em}]{Paul Suganthan}
\author[ \hspace{-0.3em}]{Rui Meng}
\author[ \hspace{-0.3em}]{Shen Gao}
\author[ \hspace{-0.3em}]{Shengcao Cao}
\author[ \hspace{-0.3em}]{Wing Lowe}
\author[ \hspace{-0.3em}]{Yi-Ting Chen}
\author[ \hspace{-0.3em}]{Yichang Chen}
\author[ \hspace{-0.3em}]{Zhongli Ding}
\author[ \hspace{-0.3em}]{Kendra Byrne}
\author[ \hspace{-0.3em}]{Cormac Brick}
\author[ \hspace{-0.3em}]{Olivier Lacombe}
\author[ \hspace{-0.3em}]{Qin Yin}
\author[ \hspace{-0.3em}]{Tris Warkentin}
\author[ \hspace{-0.3em}]{Armand Joulin}
\author[ \hspace{-0.3em}]{Tom Duerig}
\author[ \hspace{-0.3em}]{Mojtaba Seyedhosseini}

\correspondingauthor{hschechter@google.com}

\affil[ \hspace{-0.3em}]{EmbeddingGemma Team, Google\footnote{See Contributions and Acknowledgements section. $^*$Co-first authors; equal contributions.}}

\begin{abstract}
We present EmbeddingGemma~2, an open multimodal embedding model which natively represents text, images, audio, video, and combinations thereof in a unified space. Our approach introduces improvements such as simplified per-layer embeddings and a multi-stage curriculum with modality and objective progression which ensure strong quality across unimodal, cross-modal, and mixed-modal evaluations. Notably, in comprehensive embedding benchmarks across all modalities, EmbeddingGemma~2 establishes a new Pareto frontier, outperforming both specialist models and models two to twelve times larger. Furthermore, our design offers flexibility for resource-constrained environments, supporting selective modality encoder loading, adjustable vision token budgets, truncatable embedding dimensions, and quantized weights. We present ablations analyzing key decisions and release EmbeddingGemma~2 to promote further research.
\end{abstract}
\begin{document}

\maketitle
\section{Introduction}
Embedding models map data to fixed-length vector representations, placing semantically similar data near each other in the embedding space. Text embedding models have become a popular choice for tasks like information retrieval, semantic similarity, clustering, and retrieval-augmented generation. However, to work with audio and visual data, these require additional tools like automatic speech recognition (ASR) and image captioners which both add computational cost and fail to capture signals like fine-grained visual details and non-verbal or environmental sounds.

Multimodal embedding models like CLIP \citep{clip}, ALIGN \citep{align}, and SigLIP \citep{siglip} add a modality tower for vision, and so achieve strong cross-modal quality, but they are unable to natively process mixed-modal input to capture interactions between input modalities. Recent efforts have adapted Multimodal Large Language Models (MLLMs) into embedding models that allow interleaved mixed-modal input and support more complex tasks such as visual question answering and visual grounding \citep{e5v, vlm2vec, mmembed, gme}; subsequent work has built on more capable MLLMs to incorporate video and audio modalities \citep{gemembed2, omniembednemotron, lcoembed, e5omni, ovis}. However, as MLLMs have grown in size, so have embedding models adapted from them. These models typically contain several billion parameters, making them prohibitively expensive for applications that are latency-sensitive, high-throughput, or require on-device deployment for privacy or offline access.

\begin{figure}[t]
\centering
\includegraphics[width=0.75\textwidth]{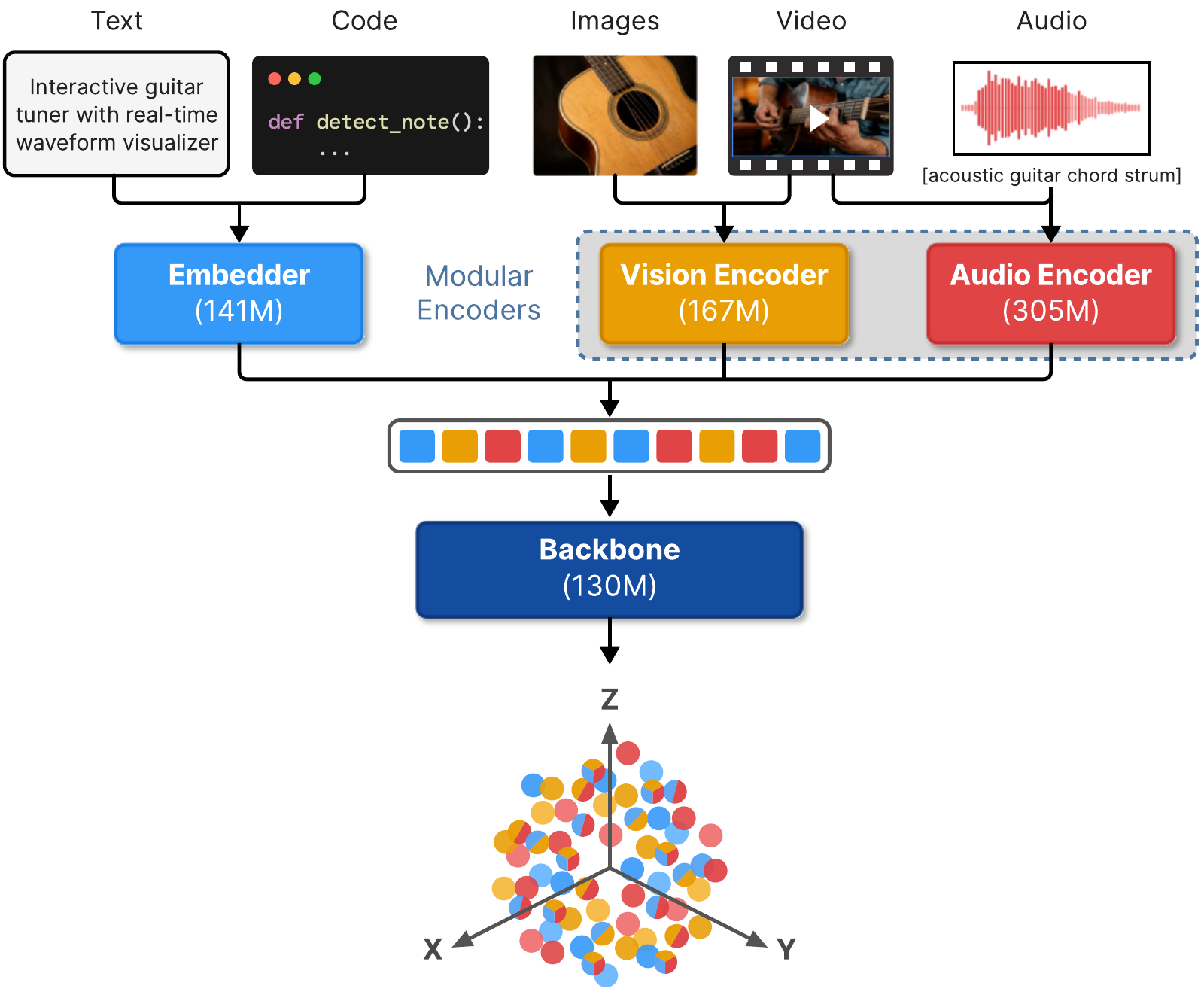}
\vspace{-0.2em}
\caption{EmbeddingGemma 2 natively maps text, images, video, audio, and combinations thereof into a unified, semantically-aligned vector space.
}
\label{fig:main}
\end{figure}

In this paper, we introduce EmbeddingGemma 2,\footnote{Our model is available at \url{https://ai.google.dev/gemma/docs/embeddinggemma}.} a 744M-parameter general-purpose multimodal embedding model that maps text, images, video, and audio---as well as arbitrary interleaved combinations thereof---into a single unified embedding space (\Cref{fig:main}). EmbeddingGemma 2 combines a 271M-parameter bidirectional transformer backbone (embedder included) with a 167M-parameter vision encoder and a 305M-parameter audio encoder. The modality encoders independently convert media into sequences of soft token embeddings that enter the shared transformer backbone, which means these parameters are modular: EmbeddingGemma 2 can alternatively be run with 271M parameters as a text-only model, with 439M as a vision-language model, or with 577M as an audio-language model.

The model builds on the architecture and weights from the capable Gemma 4 model family \citep{gemma4} and learns from the powerful Gemini Embedding 2 model \citep{gemembed2}. Our training regime consists of three stages spanning various tasks, modalities, and domains, which results in a broadly capable model, excelling at problems like code search, visual question answering, video moment retrieval, audio classification, and much more. We separate text-only and multimodal training as a form of curriculum learning, and separate distillation from contrastive and regularization objectives in order to effectively learn an embedding space tailored to the model. For extreme resource-constrained settings, the model offers high-quality truncated embeddings and separate checkpoints adapted with quantization-aware training.

We present ablation studies explaining our architectural and training choices. We find that simplifying per-layer embeddings to omit the dedicated lookup table not only reduces cost but improves performance; we present the effects of using different vision token budgets during training and inference, observing it only significantly affects tasks like document retrieval which require fine-grained visual understanding; we demonstrate the importance of co-training the vision encoder; we show the impact of finetuning with contrastive learning rather than distillation; and we confirm the value of native audio understanding over ASR.

\begin{figure}[!t]
\centering
\includegraphics[width=\textwidth]{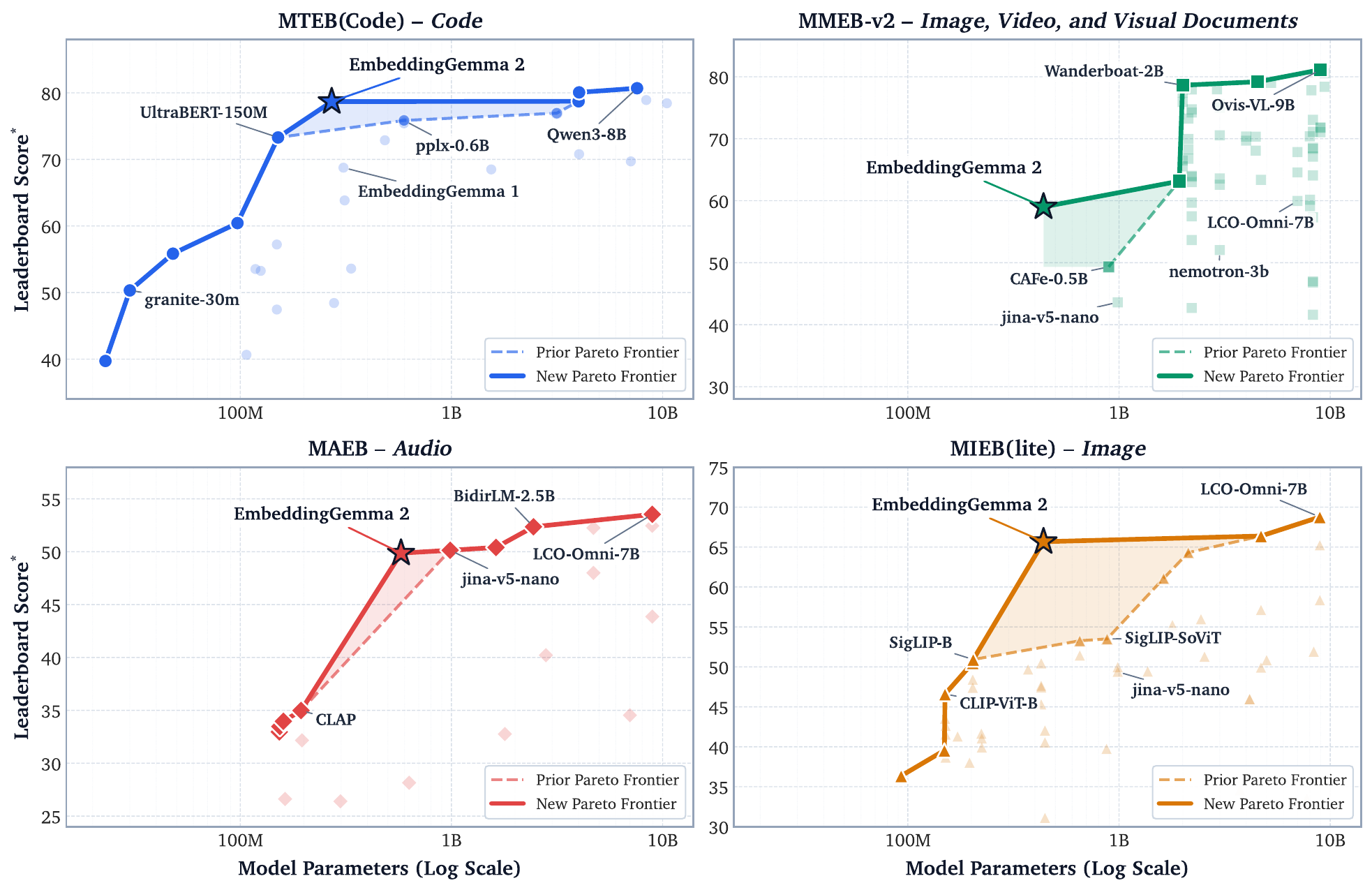}
\caption{
EmbeddingGemma 2 establishes a new Pareto frontier across benchmarks covering code, image, video, visual documents, and audio. Methods and detailed results in \Cref{sec:evaluation}.\\
$^*$Mean(Task) for MTEB(Code), MAEB, and MMEB-v2; Mean(Type) for MIEB(lite).
}
\label{fig:frontiers}
\end{figure}
Finally, we evaluate the model on seven major, comprehensive benchmark suites spanning text, code, image, video, document, speech, and sound inputs. EmbeddingGemma 2 establishes a new Pareto frontier across benchmarks spanning all modalities (\Cref{fig:frontiers}). Despite using fewer parameters for text than its predecessor and handling more modalities, it improves on both multilingual text and code benchmarks. On multimodal evals, EmbeddingGemma 2 often matches or exceeds the quality of specialized and multi-billion-parameter models.

\section{EmbeddingGemma 2}
\subsection{Architecture}

\label{sec:architecture_backbone}
\paragraph{Backbone} The backbone follows a transformer architecture \citep{transformer} with bidirectional attention. Given a tokenized input sequence $\mathbf{T}$ of length $L$, we map discrete text tokens via an embedder $\mathcal{E}$ into input token embeddings $\mathbf{T}_{\text{in}} = \mathcal{E}(\mathbf{T}) \in \mathbb{R}^{L \times d_M}$, where $d_M$ is the model dimension used for the transformer's inner representations (note multimodal inputs are incorporated into $\mathbf{T}_{\text{in}}$ as well, described in detail in \Cref{sec:input}). We then apply $\mathcal{M}_n$, an $n$-layer transformer with bidirectional attention, to produce a sequence of output token embeddings $\mathbf{T}_{\text{out}} = \mathcal{M}_n(\mathbf{T}_{\text{in}}) \in \mathbb{R}^{L \times d_M}$. Next, we employ mean pooling $\mathcal{P}$ to average the output token embeddings along the sequence axis to generate a single embedding representing all the information in the input, producing $\mathbf{P}_{\text{embed}} = \mathcal{P}(\mathbf{T}_{\text{out}}) \in \mathbb{R}^{d_M}$. Finally, we apply a randomly initialized linear projection $f$ to scale the embedding to the target dimension $d$, followed by $\ell_2$-normalization to produce the final unit-norm embedding, $\mathbf{E} = \frac{f(\mathbf{P}_{\text{embed}})}{\|f(\mathbf{P}_{\text{embed}})\|_2} \in \mathbb{R}^{d}$. In EmbeddingGemma 2, we use $n=24, d_M=512, d=768$.

Notably, the transformer $\mathcal{M}_n$ harnesses per-layer embeddings (PLE), as introduced in Gemma 3n \citep{gemma3n}. These act as residual inputs which give each layer a path to receive tailored information directly from the input token embeddings. However, instead of additionally utilizing a large, separate token lookup table specifically for PLE, EmbeddingGemma 2 simply adapts the input embeddings $\mathbf{T}_{\text{in}}$ produced by the main embedding table and modality encoders (comparison in \Cref{sec:ple_ablation}). For each layer $l \in [n]$, we project the input token embeddings into a layer-specific representation $\mathbf{P}_l = \text{RMSNorm}(p_l(\mathbf{T}_{\text{in}})) \in \mathbb{R}^{L \times d_P}$ via linear projection $p_l$. We use the layer's output $\mathbf{H}'_l \in \mathbb{R}^{L \times d_M}$ to compute gate $\mathbf{G}_l = \sigma(g_l(\mathbf{H}'_{l})) \in \mathbb{R}^{L \times d_P}$, where $g_l$ is a linear projection and $\sigma$ is a GELU activation \citep{gelu}. We then modulate the per-layer input and project it back to the model dimension via linear projection $u_l$, producing update $\mathbf{U}_l = \text{RMSNorm}(u_{l}(\mathbf{G}_l \odot \mathbf{P}_l)) \in \mathbb{R}^{L \times d_M}$, to finally yield $\mathbf{H}_l = \mathbf{H}'_l + \mathbf{U}_l$ to be processed by the next layer. In EmbeddingGemma 2, we use $d_P = d_M = 512$.  

\paragraph{Modality Encoders} As in Gemma 4, the vision encoder is a vision transformer \citep[ViT,][]{vit} with a patch size of 16 and the audio encoder consists of two downsampling convolution layers followed by twelve conformer layers \citep{conformer}, based on the Universal Speech Model (USM) architecture \citep{usm}. The vision and audio encoders produce soft tokens with dimensionalities $d_V = 768$ and $d_A = 1536$, respectively. Using learned linear projections $\phi_{\text{vis}}$ and $\phi_{\text{aud}}$, we map these to the model dimension ${d_M}$. These soft tokens are interleaved with the token embeddings at their corresponding positions to form a combined input token embedding sequence $\mathbf{T}_{\text{in}} \in \mathbb{R}^{L \times d_M}$, which is processed as described above. See \Cref{sec:input} below for more details on how multimodal inputs are handled.

\subsection{Input}
\label{sec:input}
In general, an input $x$ consists of an ordered sequence of $M \ge 1$ modality segments $x = (s_1, s_2, \dots, s_M)$, where each segment $s_m$ belongs to one of four modalities:
\begin{enumerate}[leftmargin=*]
  \setlength\itemsep{0.5em}
  \item \textbf{Text} is tokenized into $N_{\text{txt}}$ tokens which are mapped via the embedding table to a sequence of token embeddings: $\mathbf{T}_{\text{txt}} = \mathcal{E}(s_m) \in \mathbb{R}^{N_{\text{txt}} \times d_M}$.
  \item \textbf{Images} are each processed by the vision encoder $\mathcal{V}$ and spatially pooled to below a fixed budget of $N_{\text{vis}}$ soft tokens, $\mathbf{T}'_{\text{vis}} = \mathcal{V}(s_m) \in \mathbb{R}^{N_{\text{vis}} \times d_V}$. $N_{\text{vis}}$ defaults to $280$ for images and $140$ for video during training, but is configurable to $70$, $140$, $280$, $560$, or $1120$ at inference time---see how this impacts quality in \Cref{sec:budget_ablation}. These soft tokens are then projected by the vision adapter and wrapped with delimiter tokens: $\mathbf{T}_{\text{vis}} = \texttt{<|SOI|>} \oplus \phi_{\text{vis}}(\mathbf{T}'_{\text{vis}}) \oplus \texttt{<|EOI|>} \in \mathbb{R}^{(N_{\text{vis}}+2) \times d_M}$.
  \item \textbf{Audio} is processed in 40ms chunks with Mel filterbank inputs, producing 25 soft tokens per second of audio: $\mathbf{T}'_{\text{aud}} = \mathcal{A}(s_m) \in \mathbb{R}^{N_{\text{aud}} \times d_A}$. These are projected by the audio adapter and wrapped with different delimiter tokens to produce: $\mathbf{T}_{\text{aud}} = \texttt{<|SOA|>} \oplus \phi_{\text{aud}}(\mathbf{T}'_{\text{aud}}) \oplus \texttt{<|EOA|>} \in \mathbb{R}^{(N_{\text{aud}}+2) \times d_M}$.
  \item \textbf{Video} clips are typically sampled at 1 frame per second to begin. Then, for longer videos, frames are further sampled to fit a per-video frame budget $N_{\text{frm}}$---this is sometimes via truncation, by selecting the first $N_{\text{frm}}$ frames, and sometimes via uniform sampling, by selecting $N_{\text{frm}}$ evenly spaced frames over the entire video. During our training and evaluations we use $N_{\text{frm}}=32$. Since each frame is processed by the vision encoder as an image, this results in up to $N_{\text{frm}}$ segments, each with at most $N_{\text{vis}}$ soft tokens and two image delimiter tokens.
  \vspace{0.5em}
  \\Optionally, frames can be interleaved with audio tokens from the video's soundtrack, with each frame coming first and then the audio that follows it divided into 1-second segments. Usually, this corresponds to a single 1-second audio segment per frame, but if frames are truncated we might have multiple trailing audio segments, and if frames are uniformly sampled, we might have multiple audio segments between frames. In interleaved video-audio, we include text timestamps in \texttt{MM:SS} format both before each frame and before each audio segment, to indicate when the frame is taken and when the audio chunk begins. Thus, video is represented as a combination of the former modalities, yielding an embedding sequence $\mathbf{T}_{\text{vid}}$ composed of interleaved sequences of $\mathbf{T}_{\text{txt}}$, $\mathbf{T}_{\text{vis}}$, and $\mathbf{T}_{\text{aud}}$.
\end{enumerate}

Then, denoting as $\mathbf{T}_{m} \in \{\mathbf{T}_{\text{txt}}, \mathbf{T}_{\text{vis}}, \mathbf{T}_{\text{aud}}, \mathbf{T}_{\text{vid}}\}$ the embedding sequence of segment $s_m$, the final concatenated sequence is $\mathcal{T}(x) = \texttt{<|BOS|>} \oplus \mathbf{T}_{1} \oplus \mathbf{T}_{2} \oplus \dots \oplus \mathbf{T}_{M} \oplus \texttt{<|EOS|>} \in \mathbb{R}^{L \times d_M}$, with a total sequence length $L$ of up to $8192$ during training. This is consumed by the model as $\mathbf{T}_{\text{in}}$ in \Cref{sec:architecture_backbone}. 

Each training example includes multiple inputs: a query $q_i$, a positive target (or \textit{passage}) $p_i^+$, and (optionally) hard negative targets $\{p_i^-\}$. An example may also include prescribed task prefixes $t_q$ and $t_p$ for queries and targets respectively, describing the nature of the task (see \Cref{sec:task_instructions}). For instance, for retrieval we use ``\textit{task: search result | query:} \{text\}'' and ``\textit{title:} \{title | `none'\} \textit{| text:} \{text\}''. The query and targets are embedded as vectors in $\mathbb{R}^d$:
\begin{equation}
\mathbf{q}_i = f(\mathcal{P}(\mathcal{M}_n(\mathcal{T}(t_q \oplus q_i)))), \quad \mathbf{p}_i^{\pm} = f(\mathcal{P}(\mathcal{M}_n(\mathcal{T}(t_p \oplus p_i^{\pm})))).
\end{equation}
\subsection{Objective}
EmbeddingGemma 2 was trained using multiple losses. Here we introduce the loss functions themselves, and in \Cref{sec:recipe} we explain how they are introduced during training.
\paragraph{Contrastive loss} The first is a noise-contrastive estimation (NCE) loss with in-batch, hard, and same-tower negatives. Given a batch of size $B$, we define the contrastive loss as:
\begin{equation}
\begin{aligned}
\mathcal{L}_C &= -\frac{1}{B} \sum_{i=1}^{B} \log \frac{Z_i^{\text{P}}}{Z_i^{\text{P}} + Z_i^{\text{HN}} + Z_i^{\text{IB}} + Z_i^{\text{STN}}} \\[4pt]
Z_i^{\text{P}} &= e^{\text{sim}(\mathbf{q}_i, \mathbf{p}_i^+)/\tau} \\
Z_i^{\text{HN}} &= \sum_{\mathbf{p}_i^-} e^{\text{sim}(\mathbf{q}_i, \mathbf{p}_i^-)/\tau} \\
Z_i^{\text{IB}} &= \sum_{j=1}^{B} \mathbb{1}_{\text{TN}}(i, j) \, e^{\text{sim}(\mathbf{q}_i, \mathbf{p}_j^+)/\tau} \\
Z_i^{\text{STN}} &= \sum_{j=1}^{B} \mathbb{1}_{\text{TN}}(i, j) \, \bigl(e^{\text{sim}(\mathbf{q}_i, \mathbf{q}_j)/\tau} + e^{\text{sim}(\mathbf{p}_i^+, \mathbf{p}_j^+)/\tau}\bigr)
\end{aligned}
\end{equation}
where $\text{sim}(\mathbf{x}, \mathbf{y}) = \mathbf{x}^\top \mathbf{y} / \|\mathbf{x}\| \|\mathbf{y}\|$ is cosine similarity. Above, $Z_i^{\text{P}}$ corresponds to the positive, $Z_i^{\text{HN}}$ corresponds to the hard negatives, $Z_i^{\text{IB}}$  corresponds to in-batch negatives, and $Z_i^{\text{STN}}$ corresponds to same-tower negatives \citep{samtone}. $\mathbb{1}_{\text{TN}}$ masks out false negatives from duplicates:
\begin{align}
\mathbb{1}_{\text{TN}}(i, j) &=
\begin{cases}
0 & \text{if } \mathbf{q}_i = \mathbf{q}_j \text{ or } \mathbf{p}_i^+ = \mathbf{p}_j^+, \\
1 & \text{otherwise,}
\end{cases}
\end{align}

We forego the hardness weighting mechanism from our previous model as it amplifies noisy or weakly associated positive pairs, especially during early training, and compounds gradient scale imbalances between tasks of different modalities. We add same-tower negatives to prevent queries and targets from clustering into disjoint subspaces.

For listwise ranking data with $P$ ordered positives $(\mathbf{p}_{i,1}^+ \succ \dots \succ \mathbf{p}_{i,P}^+)$ per query, we train both on extracted triplets using $\mathcal{L}_C$ and on the full ranked lists using a listwise loss that extends $\mathcal{L}_C$ based on the Plackett--Luce model, as in \citet{Xia2008ListwiseAT}:
\begin{equation}
\begin{aligned}
\mathcal{L}_C^{\text{list}} &= -\frac{1}{BP} \sum_{i=1}^{B} \sum_{r=1}^{P} \log \frac{Z_{i,r}^{\text{P}}}{\sum_{k=r}^{P} Z_{i,k}^{\text{P}} + Z_i^{\text{HN}} + Z_i^{\text{IB}}} \\[4pt]
Z_{i,r}^{\text{P}} &= e^{\text{sim}(\mathbf{q}_i, \mathbf{p}_{i,r}^+)/\tau}, \qquad
Z_i^{\text{IB}} = \sum_{j=1}^{B} \mathbb{1}_{\text{TN}}(i, j) \Bigl( \sum_{k=1}^{P} e^{\text{sim}(\mathbf{q}_i, \mathbf{p}_{j,k}^+)/\tau} + \sum_{\mathbf{p}_j^-} e^{\text{sim}(\mathbf{q}_i, \mathbf{p}_j^-)/\tau} \Bigr)
\end{aligned}
\end{equation}
Here, we contrast each rank-$r$ positive against all lower-ranked positives ($k > r$) in addition to hard and in-batch negatives, with $\mathbb{1}_{\text{TN}}(i, j)$ adapted to mask out in-batch negatives that duplicate $\mathbf{p}_{i,r}^+$. We also add small squared-hinge penalties on positive and hard-negative similarities to calibrate the absolute score distribution for threshold-based tasks.

\paragraph{Spread-out regularizer} The second loss is based on the global orthogonal regularizer (GOR) introduced in \citet{spreadout}. The idea is to make the embeddings of a random pair of inputs have similar statistical properties (mean and second moment) as two points independently and uniformly sampled from the unit sphere. Following EmbeddingGemma 1 \citep{embeddinggemma}, we use only the ``second moment'' term of the regularizer's original definition, as we find this also sufficiently pushes the ``mean'' term towards its target value:
\begin{align}
\mathcal{L}_S &= \frac{1}{B\,(B-1)} \sum_{1 \le i \ne j \le B} (\mathbf{q}_{i}^\top \mathbf{q}_{j})^2 +\, \frac{1}{B\,(B-1)} \sum_{1 \le i \ne j \le B} (\mathbf{p}_{i}^{+\top} \mathbf{p}_{j}^+)^2 .
\end{align}
This loss encourages the model to produce embeddings that are spread out over the embedding space, to fully utilize its expressive power. Embedding spaces have been observed to form narrow cones \citep{mindthegap, on_sentence_embeddings, representation_degeneration, how_contextual_are}, but there are theoretical and empirical benefits to isotropy \citep{uniformity, isotropy_postprocessing}, including performance in downstream tasks. Note that pushing random pairs toward orthogonality complements the squared-hinge penalties above to make scores more calibrated and interpretable. Enforcing isotropy also aims to ensure that a) the model is robust to weight and embedding quantization, and b) the embeddings can be retrieved efficiently in vector databases using approximate nearest neighbor (ANN) algorithms.

\paragraph{Distillation loss} The final is an embedding matching loss based on \citet{kim2023embeddistillgeometricknowledgedistillation}. In contrast to previous distillation research which has relied only on the teacher's query-document relevance scores as a signal \citep{moreira2025nvretrieverimprovingtextembedding,santhanam2022colbertv2effectiveefficientretrieval}, this loss directly aligns EmbeddingGemma 2's embedding space with that of a teacher model, allowing it to more effectively learn from the larger, more powerful Gemini Embedding 2 model. Unlike \citet{kim2023embeddistillgeometricknowledgedistillation}, which applies a projection layer to distill from a model with different dimensionality, we instead make use of Gemini Embedding 2's support for lower dimensions and simply truncate the teacher embeddings. As in \citet{embeddinggemma}, we apply embedding matching not only to queries and targets but also to hard negative targets, which intuitively serves to transfer the teacher's ability to discriminate between queries and their hard negatives, and empirically does improve performance. We weigh these individual losses uniformly, letting
\begin{align}
\mathcal{L}_D &= \mathcal{L}_D^{\mathbf{q}} + \mathcal{L}_D^{\mathbf{p}^+} + \mathcal{L}_D^{\mathbf{p}^-}, \quad \text{where} \quad \mathcal{L}_D^{\mathbf{u}} = \frac{1}{B} \sum_{i=1}^{B} \bigl\| \mathbf{u}_i - \tilde{\mathbf{u}}_i \bigr\|_2^2 \quad \text{for } \mathbf{u} \in \{\mathbf{q}, \mathbf{p}^+, \mathbf{p}^-\},
\end{align}
and $\tilde{\mathbf{u}}_i$ denotes the truncated, $\ell_2$-normalized teacher embedding for input $\mathbf{u}_i$ (with $\mathcal{L}_D^{\mathbf{p}^-}$ averaged over hard negatives when multiple $\mathbf{p}_i^-$ are present).

\paragraph{Matryoshka Representation Learning} We adapt the contrastive, spread-out, and distillation losses with Matryoshka Representation Learning (MRL) \citep{kusupati2024matryoshkarepresentationlearning}, splitting each loss into $k$ separate losses applied to $k$ overlapping sub-dimensions of the embedding. EmbeddingGemma 2 provides $768$-dimensional embeddings, additionally supporting $512$-, $256$-, and $128$-dimensional embeddings via MRL.

\subsection{Recipe}
\label{sec:recipe}

\paragraph{Initialization} EmbeddingGemma 2 builds directly on Gemma 4's extensive world knowledge and capabilities. The backbone is adapted from a pretrained Gemma 4 decoder-only model, switching the causal attention mask to bidirectional attention to enable fully contextual token representations \citep{e5, gte, geminiembedding, gemembed2}. We omit explicit adaptation stages for bidirectional attention \citep{embeddinggemma, llm2vec}, which can be costly and have been shown to be unnecessary for learning expressive representations \citep{embeddinggemma, lee2025nvembedimprovedtechniquestraining}. We also remove the PLE-specific token lookup table (detailed in \Cref{sec:ple_ablation}), and add mean pooling \citep{gritlm, embeddinggemma}, a linear projection, and $\ell_2$-normalization after the transformer. The vision encoder is initialized with the smaller Gemma 4 vision encoder and further trained in the two multimodal stages, which we show in \Cref{sec:ablations} to be critical for quality. The audio encoder is bit-identical to that in the Gemma 4 E2B and E4B models \citep{gemma4} and kept frozen during training.

We then train EmbeddingGemma 2 on a mixture of embedding tasks spanning various modalities and task types. In total, the model sees approximately 2.9T tokens, of which $\sim$300B are seen during text pre-finetuning, $\sim$2T during multimodal pre-finetuning, and $\sim$600B during multimodal finetuning.

\paragraph{Text Pre-finetuning}
We begin with a pre-finetuning stage like the one in \citet{embeddinggemma}. This entails a large-scale unsupervised data mixture of query-target pairs from question answering, sentence similarity, code retrieval, and web search tasks using various languages. As before, we opt for a larger batch size both to stabilize the gradient and to provide more (in-batch) negatives for each input example. In this stage we almost exclusively employ distillation loss, with only an additional contrastive loss using a small weight. Here, as with subsequent stages, tasks are sampled at pre-determined, empirically defined rates. We found the inclusion of a text-only stage important for the model's final text quality. This stage can be seen as a form of curriculum learning, or as an implicit warm-up for the bidirectional attention, similar to \citet{llm2vec}; it has also been shown to enhance the discriminability of both text and non-text embeddings \citep{lcoembed}.

\paragraph{Multimodal Pre-finetuning}
We then continue with pre-finetuning, but with a new mixture which also includes inputs with images, audio, video, and documents across unimodal, cross-modal, and mixed-modal tasks, most with additional hard negatives. The text tasks in this stage and the next include those described in the finetuning stage of \citet{embeddinggemma}, which aim to cover task diversity, language diversity, and coding capability. In this stage, we exclusively employ distillation loss, to inherit the teacher's ability to map diverse and mixed modalities into a shared semantic space. We continue training with larger batch sizes and larger learning rates, but train for twice as many steps as in text pre-finetuning to cover the expanded multimodal training set.

\paragraph{Multimodal Finetuning}
In the finetuning stage, we completely replace distillation with contrastive loss. This allows the model to learn a tailored embedding space independent of its teacher, as the teacher's embedding space is not optimized for the student's architecture, size, and MRL sub-dimensions. We also add other losses which aim to shape the final embedding space to have desirable properties: spread-out regularizer and MRL losses. As in \citet{embeddinggemma}, we transition to smaller learning rates and batch sizes, and shift towards higher-quality, task-specific datasets, such as tool retrieval, audio captioning, video ranking, medical imaging, visual product search, cropped-object retrieval, and academic question-answering. 

\paragraph{Model Merging}
\label{sec:model_souping}
Finally, we combine checkpoints from our finetuning run by averaging their weights, to improve our final model's quality and robustness. Rather than running multiple finetuning jobs, either with different data mixture rates \citep{embeddinggemma} or with different hyperparameters \citep{souping}, we merge checkpoints along a single finetuning trajectory \citep{same_run_souping}. Specifically, we compute an unweighted average between the final checkpoint from the finetuning run and a significantly earlier checkpoint from the same run. Like \citet{wiseft}, which similarly merges pretrained and finetuned checkpoints, we found that this helps mitigate degradation on tasks that overfit during finetuning, while also further improving performance on tasks that already benefited from finetuning.

\paragraph{Quantization-Aware Training}
\begin{table}[t!]
\centering
\resizebox{1\columnwidth}{!}{
\begin{tabular}{l|l|cc|cc|c|cc|ccc}
\toprule
& & \multicolumn{2}{c|}{MTEB(Multilingual, v2)} & \multicolumn{2}{c|}{MTEB(English, v2)} & MTEB(Code) & \multicolumn{2}{c|}{MIEB(lite)$^\ddagger$} & \multicolumn{3}{c}{MSEB(Retrieval)$^\ddagger$}\\
\cmidrule(lr){3-4} \cmidrule(lr){5-6} \cmidrule(lr){7-7} \cmidrule(lr){8-9} \cmidrule(lr){10-12}
Model & Precision & Mean(Task) & Mean(Type) & Mean(Task) & Mean(Type) & Mean(Task) & Mean(Task) & Mean(Type) & Overall & In-Lang & Cross-Lang \\
\midrule
\multirow{3}{*}{EG 2} & bfloat16 & 61.36 & 53.82 & 68.46 & 63.86 & 78.68 & 60.17 & 65.67 & 69.54 & 71.63 & 67.68 \\
& int4 (Q4\_0)$^*$ & 60.96 & 53.47 & 68.48 & 63.91 & 76.85 & 56.59 & 61.95 & 64.99 & 68.56 & 61.80 \\
& int4 (Q4\_C)$^\dagger$ & 60.84 & 53.36 & 68.49 & 63.93 & 76.70 & 56.71 & 62.26 & 63.92 & 67.93 & 60.33 \\
\midrule
\multirow{2}{*}{EG 1} & bfloat16 & 61.15 & 54.31 & 69.67 & 65.11 & 68.76 & --- & --- & --- & --- & --- \\
& int4 (Q4\_0)$^*$ & 60.62 & 53.61 & 69.31 & 64.65 & 67.99 & --- & --- & --- & --- & --- \\
\bottomrule
\end{tabular}
}
\caption{Comparison of raw and quantized EmbeddingGemma 2 and EmbeddingGemma 1 checkpoints across benchmarks covering text, vision, and audio: MTEB, MIEB(lite), and MSEB(Retrieval).\\$^*$Per-block int4. $^\dagger$Per-channel int4. $^\ddagger$Different eval harnesses between float and quantized runs.
}
\label{tab:qat_quality_table}
\end{table}
We additionally provide quantized versions of EmbeddingGemma 2 in different quantization configurations. We provide checkpoints with two types of weight representations for the backbone: int4 per-block (Q4\_0) and int4 per-channel (Q4\_C). For the modality encoders, we follow the quantization schemes used in Gemma 4: the vision encoder uses 8-bit precision for both weights and activations (W8A8), and the audio encoder uses 8-bit activation precision and \{2, 4, 8\}-bit weight precision varying by layer cluster.

We apply quantization-aware training \citep[QAT,][]{qat} to both the backbone and vision encoder to minimize quality degradation (the audio encoder is taken directly from Gemma 4). However, instead of applying QAT directly in the finetuning stage as in \citet{embeddinggemma}, we introduce a separate post-merging tail-patch stage in which we distill (and initialize) our QAT checkpoint from the frozen float checkpoint \citep{apprentice}. This allows for a more resource-efficient QAT stage and narrows the gap in quality between the float and quantized checkpoints for some evaluations, notably MTEB(English, v2). \Cref{tab:qat_quality_table} shows the quality is comparable between our quantized checkpoints and our raw checkpoint (evaluated at half-precision bfloat16 weights). With Q4\_C quantization, the peak on-device memory footprint during inference (including runtime compilation overhead) on a Google Pixel 11 Pro is reduced to 191~MB for text-only, 447~MB for text and vision, and 567~MB for the full multimodal model.

\section{Ablation Studies}
\label{sec:ablations}
We analyze some key design choices to better understand how they contributed to EmbeddingGemma 2's ability to produce high-quality embeddings across modalities, domains, and task types. Baseline and ablation models use an intermediate recipe and so numbers are not representative of the final model. Unless otherwise stated, the experiments run finetuning on a pre-finetuned model. We evaluate the models on MMEB-v2 as it covers various tasks involving combinations of text, images, documents, and video, excluding only audio.

\subsection{Per-Layer Embeddings Table}
\begin{table}[t]
\centering
\resizebox{0.82\columnwidth}{!}{
\begin{tabular}{l|c|ccc}
\toprule
& \multirow{2}{*}{\parbox{2.4cm}{\centering{\textbf{MMEB-v2}\\\textbf{Overall}}}} & \multirow{2}{*}{\parbox{2.6cm}{\centering Image\\(Hit@1)}} & \multirow{2}{*}{\parbox{2.6cm}{\centering Video\\(Hit@1)}} & \multirow{2}{*}{\parbox{2.6cm}{\centering VisDoc\\(NDCG@5)}} \\
PLE Strategy & & & & \\
\midrule
Projection-only & \textcolor{blue}{\textbf{55.46}} & \textcolor{blue}{\textbf{54.67}} & \textcolor{blue}{\textbf{47.82}} & \textcolor{blue}{\textbf{62.37}} \\
\midrule
With lookup & 55.07 & 54.57 & 47.13 & 61.76 \\
\bottomrule
\end{tabular}
}
\caption{Results with and without a dedicated Per-Layer Embeddings (PLE) lookup table.}
\label{tab:ple_ablation}
\end{table}
\label{sec:ple_ablation}
Gemma 3n and Gemma 4 incorporate Per-Layer Embeddings (PLE) to inject token-level information into intermediate transformer layers. Specifically, they combine a learned projection of the main token embedding with an additional dedicated PLE lookup table. However, a dedicated PLE lookup table provides no direct signals for non-text modalities and consumes additional memory.

We utilize only the projection, as described in detail in \Cref{sec:architecture_backbone}; this way, we continue providing per-layer inputs but can omit the lookup table. We compare our approach to PLE with a lookup table in \Cref{tab:ple_ablation}. Removing the dedicated PLE table not only reduces parameter overhead, but also strengthens performance across all three categories (notably +0.69 on Video and +0.61 on VisDoc), confirming that deriving per-layer inputs solely from the shared $d_M$-dimensional embedding space improves multimodal representation quality. We also found similar improvements in text-only evaluations, where one might expect the lookup table to help the most.

\subsection{Finetuning Stage}
\begin{table}[t]
\centering
\resizebox{0.82\columnwidth}{!}{
\begin{tabular}{l|c|ccc}
\toprule
& \multirow{2}{*}{\parbox{2.4cm}{\centering{\textbf{MMEB-v2}\\\textbf{Overall}}}} & \multirow{2}{*}{\parbox{2.6cm}{\centering Image\\(Hit@1)}} & \multirow{2}{*}{\parbox{2.6cm}{\centering Video\\(Hit@1)}} & \multirow{2}{*}{\parbox{2.6cm}{\centering VisDoc\\(NDCG@5)}} \\
Training Stage & & & & \\
\midrule
Multimodal Pre-finetuning & 52.27 & 50.10 & 43.26 & 62.26 \\
\midrule
\quad $+$\ Extra Pre-finetuning & 52.28 & 49.74 & 43.86 & \textcolor{blue}{\textbf{62.40}} \\
\quad $+$\ Finetuning Stage & \textcolor{blue}{\textbf{55.46}} & \textcolor{blue}{\textbf{54.67}} & \textcolor{blue}{\textbf{47.82}} & 62.37 \\
\bottomrule
\end{tabular}
}
\caption{Results of continued distillation versus adding a finetuning stage with contrastive learning.}
\label{tab:contrastive_ablation}
\end{table}
We investigate whether the gains from our finetuning stage are truly driven by a difference in recipe or merely by extending training. Starting from the multimodal pre-finetuning checkpoint, we compare running the multimodal finetuning stage against continuing distillation for an approximately compute-equivalent duration (the same number of steps, but with larger batch sizes on distillation). \Cref{tab:contrastive_ablation} shows that while distillation plateaus, transitioning to the finetuning recipe with contrastive and spread-out objectives boosts performance significantly for image and video tasks while yielding similar results for visual documents. This demonstrates that the finetuning stage enables the model to move beyond imitating its teacher and learn a better-optimized representation space.

\subsection{Vision Encoder Training}
\begin{table}[t]
\centering
\resizebox{0.82\columnwidth}{!}{
\begin{tabular}{l|c|ccc}
\toprule
& \multirow{2}{*}{\parbox{2.4cm}{\centering{\textbf{MMEB-v2}\\\textbf{Overall}}}} & \multirow{2}{*}{\parbox{2.6cm}{\centering Image\\(Hit@1)}} & \multirow{2}{*}{\parbox{2.6cm}{\centering Video\\(Hit@1)}} & \multirow{2}{*}{\parbox{2.6cm}{\centering VisDoc\\(NDCG@5)}} \\
Vision Encoder & & & & \\
\midrule
Trainable & \textcolor{blue}{\textbf{52.27}} & \textcolor{blue}{\textbf{50.10}} & \textcolor{blue}{\textbf{43.26}} & \textcolor{blue}{\textbf{62.26}} \\
\midrule
Frozen & 48.18 & 44.35 & 39.24 & 60.59 \\
\bottomrule
\end{tabular}
}
\caption{Results with trainable and frozen vision encoder.}
\label{tab:frozen_ablation}
\end{table}
MLLMs often keep modality encoders frozen throughout training \citep{prismatic, llava, palmE, gemma4}. This strategy is also popular in recent multimodal embedding models \citep{uembed, omniembednemotron, jinav5omni, e5v}. For MLLMs, ablations investigating whether freezing or unfreezing the vision encoder is better are mixed \citep{cambrian, prismatic, mm1}. However, we show in \Cref{tab:frozen_ablation} that, though the Gemma 4 vision encoder is quite strong by itself, training the vision encoder greatly improves performance of the resulting embeddings on multimodal tasks. Analogous gains are likely possible by further training the audio encoder. Note that both runs here start with a text pre-finetuning checkpoint and run multimodal pre-finetuning, to ensure the vision encoder has not already been updated prior to either experiment.

Some MLLMs manage to additionally keep the backbone frozen \citep{blip2}. \citet{jinav5omni} similarly joins a backbone text embedding model with modality encoders, keeping each of them frozen. This prevents users from having to backfill embedding databases created with the text embedding model; however, it is possible to translate embeddings from one model's space to another \citep{mkp}, even without access to paired data or to the original models \citep{universalgeometry}. Moreover, freezing a text-only backbone empirically seems to limit its ability to understand multimodal inputs, and indeed many of the strongest multimodal embedding models finetune the backbone alongside the modality encoders \citep{ovis, lcoembed}.

\subsection{Vision Token Budgets}
\begin{table}[t!]
\centering
\resizebox{0.95\columnwidth}{!}{
\begin{tabular}{l|cc|c|ccc}
\toprule
& \multicolumn{2}{c|}{Vision Token Budget} & \multirow{2}{*}{\parbox{2.4cm}{\centering{\textbf{MMEB-v2}\\\textbf{Overall}}}} & \multirow{2}{*}{\parbox{2.6cm}{\centering Image\\(Hit@1)}} & \multirow{2}{*}{\parbox{2.6cm}{\centering Video\\(Hit@1)}} & \multirow{2}{*}{\parbox{2.6cm}{\centering VisDoc\\(NDCG@5)}} \\
\cmidrule(lr){2-3}
Experiment & Train (Img / Vid) & Eval & & & & \\
\midrule
\multirow{4}{*}{{Inference-Time Scaling}}
& 280 / 140 & \phantom{0}70  & 50.07 & 50.90 & 46.33 & 51.61 \\
& 280 / 140 & 140 & 55.37 & 53.58 & 47.99 & 63.60 \\
& 280 / 140 & 280 & 56.15 & 55.15 & \textcolor{blue}{\textbf{48.51}} & 63.38 \\
& 280 / 140 & 560 & \textcolor{blue}{\textbf{57.02}} & \textcolor{blue}{\textbf{55.21}} & 46.52 & \textcolor{blue}{\textbf{67.62}} \\
\midrule
\multirow{4}{*}{{Train-Time Scaling}}
& 280 / \phantom{0}70 & \phantom{0}70 & 49.62 & 50.56 & \textcolor{blue}{\textbf{46.60}} & 50.49 \\
& 280 / 140 & \phantom{0}70 & \textcolor{blue}{\textbf{49.76}} & \textcolor{blue}{\textbf{50.76}} & 45.79 & \textcolor{blue}{\textbf{51.25}} \\
\cmidrule(lr){2-7}
& 280 / \phantom{0}70 & 140 & 54.66 & 53.29 & 47.16 & 62.33 \\
& 280 / 140 & 140 & \textcolor{blue}{\textbf{55.46}} & \textcolor{blue}{\textbf{54.67}} & \textcolor{blue}{\textbf{47.82}} & \textcolor{blue}{\textbf{62.37}} \\
\bottomrule
\end{tabular}
}
\caption{Results with different vision token budgets (max soft tokens per image or video frame) at train and inference time. Evaluation uses a single budget for all tasks. \textbf{Top:} Inference-time scaling of a model trained with 280 tokens per image and 140 per video frame. \textbf{Bottom:} Train-time scaling with 70 vs.\ 140 tokens per video frame. The two experiments use different intermediate recipes.}
\label{tab:vision_budget_ablation}
\end{table}
\label{sec:budget_ablation}
We train EmbeddingGemma 2 with a budget of 280 vision tokens for images and 140 tokens per frame for videos, but the underlying Gemma 4 vision encoder supports budgets from 70 to 1120 tokens. In \Cref{tab:vision_budget_ablation} (top), we show the effect of using different vision token budgets at inference time using an intermediate checkpoint trained with 280 tokens per image and 140 per frame. Overall performance scales monotonically, with some distinct modality-specific behaviors. Visual document tasks are highly sensitive to fine-grained details and continue to improve greatly past the training budget; image quality in aggregate begins to saturate at the 280-token budget (despite continued gains on several individual tasks); and video performance actually degrades between 280 tokens and 560, likely because the excessive sequence lengths dilute attention and mean-pooled representations.
 
Additionally, we compare models trained with 70 vs.\ 140 per-frame video tokens (\Cref{tab:vision_budget_ablation}, bottom) to see whether training with a smaller vision budget could improve quality when using smaller vision budgets at inference time, for resource-constrained deployment. When evaluated with 70 tokens, training with 70 video tokens improves performance on video tasks, as the model is directly exposed to 70-token frames during training. But this advantage does not hold for image or visual document tasks, which drop in quality. This is likely because a) single-image tasks, which continue to be trained with 280 tokens, are treated as a separate modality and do not benefit from seeing 70-token video frames; and b) coarser video representations discard spatial details that otherwise transfer to stronger general visual representations. The latter is further supported when evaluating with 140 tokens, where training with 70 video tokens degrades performance even more, and across modalities---this makes the 140-token video budget preferable across both evaluation regimes. More gains are likely possible by increasing train-time budgets, but with seemingly diminishing returns and at a significant computational cost.

\section{Evaluation}
\label{sec:evaluation}
We evaluate EmbeddingGemma 2 on a comprehensive set of benchmarks across diverse modalities, task types, domains, and languages. Specifically, we report numbers from the Massive Text Embedding Benchmark \citep[MTEB,][]{enevoldsen2025mmteb,muennighoff2023mtebmassivetextembedding} for natural language and code; the Massive Multimodal Embedding Benchmark v2 \citep[MMEB-v2,][]{vlm2vec2} and Massive Image Embedding Benchmark \citep[MIEB(lite),][]{mieb} for image, video, and visual documents; and the Massive Sound Embedding Benchmark and Massive Audio Embedding Benchmark for speech, music, and environmental sounds \citep[MSEB and MAEB,][]{mseb, maeb}. We provide a summary of our results in \Cref{tab:main_table} and dive into each benchmark in more detail below.
\newcommand{\rotheader}[1]{\rotatebox{65}{\makecell[l]{#1}}}
\newcommand{\mcrot}[4]{\multicolumn{#1}{#2}{\rlap{\rotatebox{#3}{#4}~}}}
\newcommand{\rotangle}{45}

\begin{table}[!t]
\centering
\resizebox{0.95\columnwidth}{!}{
\begin{tabular}{llcc|ccccc}
\Xcline{1-9}{0.08em}
\rule{0pt}{12.5ex} &
& \mcrot{1}{l}{\rotangle}{\shortstack[l]{Embedding\\Gemma 2}} 
& \mcrot{1}{l}{\rotangle}{\shortstack[l]{Embedding\\Gemma 1}} 
& \mcrot{1}{l}{\rotangle}{\shortstack[l]{Jina Embed\\v5 Omni Nano}}
& \mcrot{1}{l}{\rotangle}{\shortstack[l]{Qwen3\\Embed 0.6B}}
& \mcrot{1}{l}{\rotangle}{\shortstack[l]{InterestFM\\UIR CAFe 0.5B}}
& \mcrot{1}{l}{\rotangle}{\shortstack[l]{SigLIP\\SoViT 400M}}
& \mcrot{1}{l}{\rotangle}{\shortstack[l]{LAION CLAP\\General}} \\
\cmidrule{1-9}
\textbf{Total Parameters} & & 744M & 308M & 986M & 596M & 894M & 878M & 194M \\
\cmidrule{1-9}
\textbf{Embedding Dimension} & & 768 & 768 & 768 & 1024 & 896 & 1152 & 512 \\
\cmidrule{1-9}
\multirow{2}{3.7cm}{\textbf{MTEB(Multi, v2)} \footnotesize{\citep{enevoldsen2025mmteb}}} & Mean (Task) & 61.36 & 61.15 & \textbf{65.52} & 64.34 & -- & -- & -- \\
& Mean (Type) & 53.82 & 54.31 & \textbf{57.66} & 56.01 & -- & -- & -- \\
\cmidrule{1-9}
\multirow{2}{3.7cm}{\textbf{MTEB(Eng, v2)} \footnotesize{\citep{enevoldsen2025mmteb}}}
& Mean (Task) & 68.46 & 69.67 & \textbf{71.11} & 70.47 & -- & -- & -- \\
& Mean (Type) & 63.86 & 65.11 & \textbf{65.22} & 64.72 & -- & -- & -- \\
\cmidrule{1-9}
\multirow{2}{3.7cm}{\textbf{MTEB(Code)} \footnotesize{\citep{enevoldsen2025mmteb}}}
& \multirow{2}{*}{Mean (Task)} & \multirow{2}{*}{\textcolor{blue}{\textbf{78.68}}} & \multirow{2}{*}{68.76} & \multirow{2}{*}{--} & \multirow{2}{*}{75.42} & \multirow{2}{*}{--} & \multirow{2}{*}{--} & \multirow{2}{*}{--} \\
& & & & & & & & \\
\cmidrule{1-9}
\multirow{4}{3.7cm}{\textbf{MMEB-v2} \footnotesize{\citep{vlm2vec2}}}
& Mean (Task) & \textcolor{blue}{\textbf{59.01}} & $*$ & 43.61 & $*$ & 49.33 & -- & $*$ \\
\cmidrule{2-9}
& - Image  & \textcolor{blue}{\textbf{57.28}} & $*$ & 31.56 & $*$ & 55.43 & -- & $*$ \\
& - Video  & \textcolor{blue}{\textbf{50.67}} & $*$ & 30.73 & $*$ & 35.87 & -- & $*$ \\
& - VisDoc & 67.84 & $*$ & \textbf{71.34} & $*$ & 50.26 & -- & $*$ \\
\cmidrule{1-9}
\multirow{2}{3.7cm}{\textbf{MIEB(lite)} \footnotesize{\citep{mieb}}}
& Mean (Task) & \textcolor{blue}{\textbf{60.17}} & $*$ & 44.24 & $*$ & -- & 52.28 & $*$ \\
& Mean (Type) & \textcolor{blue}{\textbf{65.67}} & $*$ & 49.38 & $*$ & -- & 53.50 & $*$ \\
\cmidrule{1-9}
\multirow{2}{3.7cm}{\textbf{MAEB} \footnotesize{\citep{maeb}}}
& Mean (Task) & 49.87 & $*$ & \textbf{50.14} & $*$ & $*$ & $*$ & 34.98 \\
& Mean (Type) & 48.65 & $*$ & \textbf{55.31} & $*$ & $*$ & $*$ & 41.08 \\
\cmidrule{1-9}
\multirow{2}{3.7cm}{\textbf{MSEB(Retrieval)} \footnotesize{\citep{mseb}}}
& \multirow{2}{*}{Mean (Task)} & \multirow{2}{*}{\textcolor{blue}{\textbf{69.54}}} & \multirow{2}{*}{50.95$^\dagger$} & \multirow{2}{*}{--} & \multirow{2}{*}{$*$}& \multirow{2}{*}{$*$} & \multirow{2}{*}{$*$} & \multirow{2}{*}{--} \\
& & & & & & & & \\
\Xcline{1-9}{0.08em}
\end{tabular}
}
\caption{Comparison on benchmarks against a slightly larger multimodal peer and modality specialists.\\$*$\hspace{0.3em}Modality not supported. --\hspace{0.3em}Score unavailable. $^\dagger$Computed by transcribing audio with ASR.}
\label{tab:main_table}
\end{table}

\subsection{Methods}
We select as baselines the top-ranked models on each benchmark, including both modality specialists and general-purpose embedding models which support all evaluated modalities. For MTEB, MAEB, and MIEB(lite), we compare against models under 1B parameters; for MMEB-v2, we include models under 2B parameters, as small models are scarce. Because it is one of the most directly comparable models, we evaluate \texttt{jina-embeddings-v5-omni-nano} on MMEB-v2 using its official implementation and verify that, on the subset reported in \citet{jinav5omni}, our reproduced score is slightly higher. All other baseline scores are sourced from public leaderboards. We restrict our analysis to single-vector models, given the disparity in storage and latency costs with multi-vector models, and omit models trained on $>$25\% of the evaluation data, as data contamination undermines the evaluation's validity.

We evaluate EmbeddingGemma 2 using half-precision weights (bfloat16) and the seven types of task instructions detailed in \Cref{sec:task_instructions}. We use different sequence lengths depending on the modality and task type. For visual inputs, we use up to $N_{\text{frm}}=32$ frames per video at 140 vision tokens per frame and 280 vision tokens per image to reflect realistic model use, though we note that higher vision token budgets could yield further gains, particularly for visual documents (\Cref{sec:budget_ablation}). Because most baseline models support limited modalities, we count only the parameters EmbeddingGemma 2 utilized for each benchmark; other models' parameter counts are as reported on respective leaderboards.

\subsection{Text}
The MTEB suite consists of a massive collection of individual, quality-controlled evaluation tasks. Together, the 162 tasks cover 250+ natural languages and 14+ programming languages, 20 domains (e.g. legal, medical, news, programming, web), and 10 task types: bitext mining, classification, clustering, instruction retrieval, multilabel classification, pair classification, reranking, retrieval, semantic text similarity (STS), and summarization.

Despite having fewer text parameters and sharing its representation space with vision and audio, EmbeddingGemma 2 improves over its predecessor on multilingual text (driven by gains in bitext mining and classification). On English, the model sees a modest regression. Only two smaller models surpass it on multilingual text, and one on English, but they are all trained as specialist text models (\Cref{sec:appendix_leaderboards}). The model presents competitive text scores which outperform much larger models.

On code (\Cref{tab:leaderboard_code}), EmbeddingGemma 2's performance is stellar, establishing a new Pareto frontier. It outperforms every sub-4B model on the leaderboard and even comes within 2 points of the 28$\times$ larger frontrunner. In all three benchmarks, the model's truncated representations degrade minimally, especially down to 256 dimensions.

\begin{table}[t!]
\centering
\resizebox{1\columnwidth}{!}{%
\begin{tabular}{lccc|cccccccccccc}
\toprule
Model Name & Params & Dim. & \parbox{1.2cm}{\centering{\textbf{Mean}\\\textbf{(Task)}}} & {AppsR.} & {COIR} & {CESR} & {CFMT} & {CSMT} & {CSNCCR} & {CSNR} & {CTOC} & {CTODL} & {CQA} & {SOQA} & {ST2SQL} \\
\midrule
{EmbeddingGemma 2} & 271M & 768 & \textcolor{blue}{\textbf{78.7}} & \textcolor{blue}{\textbf{89.2}} & \textcolor{blue}{\textbf{86.9}} & \textcolor{blue}{\textbf{80.5}} & 85.6 & 82.2 & \textcolor{blue}{\textbf{93.8}} & 90.9 & 84.6 & 35.6 & \textcolor{blue}{\textbf{53.3}} & 89.1 & 72.5 \\
\hphantom{EmbeddingGemma 2} &  & 512 & 77.2 & \textcolor{blue}{\textbf{89.2}} & 86.6 & 79.9 & 85.3 & 82.0 & 93.7 & 90.7 & 82.9 & 35.7 & 52.7 & 87.0 & 61.1 \\
\hphantom{EmbeddingGemma 2} &  & 256 & 76.2 & 88.3 & 85.5 & 77.9 & 84.1 & 80.5 & 92.9 & 90.2 & 82.8 & 35.8 & 50.5 & 84.9 & 60.9 \\
\hphantom{EmbeddingGemma 2} &  & 128 & 71.4 & 82.5 & 80.8 & 70.0 & 78.5 & 74.8 & 89.5 & 88.1 & 77.4 & 34.6 & 44.1 & 77.0 & 59.7 \\
\midrule
pplx-embed-v1-0.6b & 596M & 1024 & 75.8 & 78.7 & 82.4 & 76.4 & 83.0 & 84.7 & 88.1 & 89.9 & 85.0 & 35.2 & 43.3 & 90.7 & 73.0 \\
UltraBERT-Embed-150M & 151M & 768 & 73.3 & 68.5 & 82.2 & 62.2 & 84.4 & 86.4 & 93.7 & 90.0 & \textbf{88.1} & 37.2 & 36.3 & \textbf{92.4} & 58.0 \\
Giga-Embeddings-instruct & 484M & 1024 & 72.9 & 81.6 & 71.8 & 61.8 & \textbf{91.2} & \textbf{89.6} & 73.9 & 87.3 & 86.5 & \textbf{37.6} & 37.9 & 92.3 & 62.9 \\
Qwen3-Embedding-0.6B & 596M & 1024 & 75.4 & 75.3 & 84.7 & 64.4 & 90.8 & 86.4 & 91.7 & \textbf{91.0} & 86.0 & 31.4 & 36.5 & 90.0 & \textbf{76.7} \\
embeddinggemma-300m & 308M & 768 & 68.8 & 84.4 & 75.5 & 62.1 & 51.4 & 80.3 & 73.7 & 90.1 & 85.5 & 33.5 & 43.6 & 86.5 & 58.4 \\
granite-embed-311m-multi & 312M & 768 & 63.8 & 60.6 & 77.6 & 56.2 & 57.0 & 77.3 & 55.5 & 86.9 & 83.2 & 34.4 & 35.4 & 86.3 & 55.7 \\
granite-embed-english-r2 & 149M & 768 & 57.2 & 14.0 & 64.7 & 59.3 & 52.5 & 77.2 & 47.7 & 80.8 & 77.1 & 35.0 & 37.0 & 91.8 & 49.6 \\
granite-embed-97m-multi & 97M & 384 & 60.4 & 50.2 & 75.4 & 53.3 & 55.2 & 75.2 & 51.9 & 85.3 & 76.5 & 34.0 & 34.0 & 82.0 & 52.3 \\
granite-embed-small-eng & 48M & 384 & 55.8 & 13.5 & 60.5 & 57.2 & 52.2 & 76.9 & 48.4 & 78.3 & 77.6 & 33.6 & 35.6 & 90.0 & 46.3 \\
granite-embed-125m-eng & 125M & 768 & 53.3 & 11.8 & 55.1 & 58.8 & 42.1 & 75.3 & 47.6 & 76.9 & 66.7 & 29.6 & 36.6 & 89.8 & 48.7 \\
b1ade-embed & 335M & 1024 & 53.6 & 6.7 & 79.8 & 55.8 & 37.8 & 72.5 & 51.8 & 86.7 & 57.0 & 22.3 & 35.3 & 83.5 & 54.0 \\
\bottomrule
\end{tabular}
}
\caption{Performance of top-ranking models under 1B parameters on MTEB(Code).}
\label{tab:leaderboard_code}
\end{table}

\subsection{Vision}
MMEB-v2 spans 78 tasks across image, video, and visual document modalities, the latter of which includes PDFs, slides, and charts. These include retrieval, classification, question answering, visual grounding, and moment retrieval. MIEB(lite) complements this with linear-probe classification, clustering, multiple choice, compositionality, and visual semantic text similarity. They both implicitly evaluate countless visual skills like object detection, counting, optical character recognition, and temporal reasoning. They also cover various permutations of cross-modal and mixed-modal situations.

EmbeddingGemma 2 is the smallest model on the MMEB-v2 leaderboard (\Cref{tab:leaderboard_multi}). It excels especially at classification and retrieval, but provides strong scores across the board---its overall score is exceeded only by models around 2B parameters. It also outperforms various recent models including vision-language specialists with more than 8B parameters. On MIEB(lite), only two models beat EmbeddingGemma 2, and they have well over 4B parameters. Compared to top models under 1B parameters (\Cref{tab:leaderboard_image}), EmbeddingGemma 2 sets the state of the art on both aggregate metrics by a wide margin (+7.9 Task, +12.2 Type). It also offers remarkable improvements in nearly all task types, notably multilingual visual STS (+22.8) and multilingual retrieval (+12.5). Even at 128 dimensions, the model outperforms every other sub-1B model at full dimensionality.

\subsection{Audio}
\begin{table}[!t]
\centering
\resizebox{1\columnwidth}{!}{%
\begin{tabular}{lccc|ccccc|ccccc|ccccc}
\toprule
& & & \raisebox{-3.9pt}[0pt][0pt]{\textbf{Mean}} & \multicolumn{5}{c|}{\textbf{Image}} & \multicolumn{5}{c|}{\textbf{Video}} & \multicolumn{5}{c}{\textbf{VisDoc}} \\
\cmidrule(lr){5-9} \cmidrule(lr){10-14} \cmidrule(lr){15-19}
Model Name & Params & Dim. & \textbf{(Task)} & CLS & QA & RET & GD & \textbf{All} & CLS & QA & RET & MRET & \textbf{All} & VDRv1 & VDRv2 & VR & OOD & \textbf{All} \\
\midrule
{EmbeddingGemma 2} & 439M & 768 & 59.0 & 54.4 & 41.9 & \textcolor{blue}{\textbf{65.8}} & 77.4 & 57.3 & \textcolor{blue}{\textbf{56.2}} & 48.6 & \textcolor{blue}{\textbf{47.2}} & \textcolor{blue}{\textbf{50.6}} & \textcolor{blue}{\textbf{50.7}} & 70.6 & \textcolor{blue}{\textbf{58.3}} & 72.7 & 63.3 & 67.8 \\
\hphantom{EmbeddingGemma 2} &  & 512 & 58.4 & 54.2 & 41.4 & 65.4 & 77.1 & 56.9 & 56.0 & 48.5 & 46.5 & 50.3 & 50.3 & 69.2 & 56.3 & 72.0 & 62.2 & 66.6 \\
\hphantom{EmbeddingGemma 2} &  & 256 & 56.2 & 53.2 & 39.5 & 63.6 & 76.5 & 55.5 & 54.6 & 48.4 & 44.9 & 49.2 & 49.3 & 64.9 & 53.2 & 67.2 & 59.5 & 62.6 \\
\hphantom{EmbeddingGemma 2} &  & 128 & 45.6 & 39.9 & 29.6 & 54.7 & 72.8 & 45.6 & 40.8 & 44.4 & 34.9 & 43.6 & 40.6 & 51.9 & 39.2 & 52.6 & 48.8 & 49.5 \\
\midrule
Eddy-vl\_embed\_1.9B\_v1 & 1.93B & 2048 & \textbf{63.2} & 54.9 & \textbf{61.7} & 64.8 & \textbf{84.7} & \textbf{63.4} & 51.5 & \textbf{57.9} & 41.9 & 48.0 & 50.0 & \textbf{78.8} & 55.7 & \textbf{79.1} & \textbf{64.9} & \textbf{72.7} \\
interestFM-UIR-CAFe & 894M & 896 & 49.3 & \textbf{56.4} & 45.3 & 57.5 & 72.0 & 55.4 & 33.9 & 41.7 & 29.7 & 39.7 & 35.9 & 56.9 & 32.6 & 68.6 & 23.8 & 50.3 \\
jina-embed-v5-omni-nano & 986M & 768 & 43.6 & 34.6 & 16.0 & 37.4 & 45.5 & 31.6 & 28.1 & 41.5 & 18.0 & 38.3 & 30.7 & 77.6 & 53.9 & 78.6 & 62.2 & 71.3 \\
\bottomrule
\end{tabular}
}
\caption{Performance of top-ranking models under 2B parameters on MMEB-v2.}
\label{tab:leaderboard_multi}
\end{table}
\begin{table}[!t]
\centering
\resizebox{\columnwidth}{!}{%
\begin{tabular}{lcccc|cccccccccc}
\toprule
& & & \multirow{2}{*}{\parbox{1cm}{\centering{\textbf{Mean}\\\textbf{(Task)}}}} & \multirow{2}{*}{\parbox{1cm}{\centering \textbf{Mean}\\\textbf{(Type)}}} & \multirow{2}{*}{\parbox{1cm}{\centering Multi.\\Retr.}} & \multirow{2}{*}{\parbox{1cm}{\centering A2A\\Retr.}} & & \multirow{2}{*}{\parbox{1cm}{\centering Doc.\\Und.}} & \multirow{2}{*}{\parbox{1cm}{\centering Image\\Class.}} & \multirow{2}{*}{\parbox{1cm}{\centering Image\\Clus.}} & \multirow{2}{*}{\parbox{1cm}{\centering Vision\\QA}} & \multirow{2}{*}{\parbox{1cm}{\centering VSTS\\(eng)}} & \multirow{2}{*}{\parbox{1cm}{\centering VSTS\\(multi)}} & \multirow{2}{*}{\parbox{1.04cm}{\centering 0-shot\\Class.}} \\
Model Name & Params & Dim. & & & & & {Comp.} & & & & & & & \\
\midrule
{EmbeddingGemma 2} & 439M & 768 & \textcolor{blue}{\textbf{60.2}} & \textcolor{blue}{\textbf{65.7}} & \textcolor{blue}{\textbf{65.4}} & \textcolor{blue}{\textbf{38.3}} & \textcolor{blue}{\textbf{45.7}} & 67.3 & 74.3 & \textcolor{blue}{\textbf{81.3}} & \textcolor{blue}{\textbf{59.1}} & 81.5 & \textcolor{blue}{\textbf{76.8}} & 67.0 \\
\hphantom{EmbeddingGemma 2} &  & 512 & 59.9 & 65.3 & 64.8 & 38.1 & \textcolor{blue}{\textbf{45.7}} & 66.5 & 74.2 & 80.0 & 58.8 & 81.6 & \textcolor{blue}{\textbf{76.8}} & 66.9 \\
\hphantom{EmbeddingGemma 2} &  & 256 & 58.6 & 64.2 & 62.9 & 37.1 & 44.5 & 64.0 & 73.7 & 78.9 & 56.4 & \textcolor{blue}{\textbf{81.8}} & \textcolor{blue}{\textbf{76.8}} & 65.5 \\
\hphantom{EmbeddingGemma 2} &  & 128 & 54.2 & 60.0 & 50.7 & 33.0 & 40.1 & 56.6 & 71.5 & 79.3 & 56.7 & 80.8 & 74.8 & 56.4 \\
\midrule
siglip-so400m-patch14-384 & 878M & 1152 & 52.3 & 53.5 & 35.4 & 32.4 & 32.8 & 46.9 & \textbf{78.8} & 75.9 & 48.0 & 69.6 & 41.4 & \textbf{73.8} \\
siglip-large-patch16-384 & 652M & 1024 & 51.4 & 53.3 & 43.5 & 31.9 & 32.1 & 44.9 & 77.7 & 75.2 & 46.8 & 69.6 & 39.8 & 71.3 \\
siglip-base-patch16-512 & 204M & 768 & 49.6 & 50.9 & 34.8 & 30.8 & 29.7 & 42.6 & 74.6 & 69.7 & 55.5 & 67.1 & 38.1 & 66.3 \\
CLIP-ViT-L-14-DataComp.XL & 428M & 768 & 49.0 & 50.4 & 22.6 & 31.0 & 31.7 & 30.8 & 75.3 & 80.4 & 54.9 & 72.5 & 35.8 & 69.4 \\
CLIP-ViT-H-14-laion2B & 986M & 1024 & 49.3 & 50.0 & 23.9 & 32.8 & 34.8 & 33.7 & 76.8 & 79.3 & 46.8 & 68.3 & 33.9 & 69.4 \\
siglip-base-patch16-256-multi & 371M & 768 & 46.6 & 49.7 & 52.9 & 28.2 & 30.7 & 22.9 & 73.4 & 68.2 & 53.3 & 63.7 & 40.3 & 63.2 \\
jina-embed-v5-omni-nano & 986M & 768 & 44.2 & 49.4 & 38.7 & 23.6 & 29.6 & \textbf{74.2} & 63.7 & 60.5 & 50.0 & 77.9 & 54.0 & 21.8 \\
CLIP-ViT-L-14-laion2B & 428M & 768 & 47.4 & 47.6 & 21.4 & 31.1 & 33.6 & 28.7 & 75.9 & 76.4 & 46.9 & 68.7 & 26.0 & 67.8 \\
clip-vit-large-patch14 & 428M & 768 & 46.4 & 47.4 & 19.8 & 26.7 & 39.4 & 29.4 & 74.5 & 71.3 & 44.8 & 69.4 & 35.1 & 63.8 \\
siglip-base-patch16-224 & 203M & 768 & 46.1 & 47.4 & 33.6 & 29.3 & 32.5 & 20.9 & 73.5 & 68.4 & 53.0 & 64.2 & 33.5 & 65.0 \\
CLIP-ViT-B-16-DataComp.XL & 150M & 512 & 45.5 & 46.6 & 19.9 & 28.3 & 31.4 & 22.7 & 73.2 & 73.6 & 56.9 & 69.6 & 28.5 & 61.9 \\
\bottomrule
\end{tabular}
}
\caption{Performance of top-ranking models under 1B parameters on MIEB(lite).}
\label{tab:leaderboard_image}
\end{table}
\begin{table}[!t]
\centering
\resizebox{\columnwidth}{!}{%
\begin{tabular}{lcccc|ccccccc}
\toprule
& & & \multirow{2}{*}{\parbox{1cm}{\centering{\textbf{Mean}\\\textbf{(Task)}}}} & \multirow{2}{*}{\parbox{1cm}{\centering \textbf{Mean}\\\textbf{(Type)}}} & \multirow{2}{*}{\parbox{1cm}{\centering A2A\\Retr.}} & \multirow{2}{*}{\parbox{1cm}{\centering Audio\\Class.}} & \multirow{2}{*}{\parbox{1cm}{\centering Audio\\Clus.}} & \multirow{2}{*}{\parbox{1cm}{\centering Multi.\\Class.}} & \multirow{2}{*}{\parbox{1cm}{\centering Pair\\Class.}} & \multirow{2}{*}{\parbox{1cm}{\centering Audio\\Rerank.}} & \multirow{2}{*}{\parbox{1.04cm}{\centering 0-shot\\Class.}} \\
Model Name & Params & Dim. & & & & & & & & & \\
\midrule
{EmbeddingGemma 2} & 577M & 768 & 49.9 & 48.7 & \textcolor{blue}{\textbf{53.6}} & \textcolor{blue}{\textbf{52.9}} & 4.1 & 38.7 & \textcolor{blue}{\textbf{67.4}} & 68.5 & 55.4 \\
\hphantom{EmbeddingGemma 2} &  & 512 & 49.8 & 48.6 & 53.4 & 52.7 & 4.3 & 38.8 & \textcolor{blue}{\textbf{67.4}} & 68.8 & 55.0 \\
\hphantom{EmbeddingGemma 2} &  & 256 & 49.1 & 48.2 & 52.1 & 52.2 & 4.2 & 38.4 & 67.2 & 68.7 & 54.8 \\
\hphantom{EmbeddingGemma 2} &  & 128 & 47.3 & 46.3 & 49.0 & 50.4 & 4.5 & 30.8 & 66.9 & 67.1 & 55.1 \\
\midrule
larger\_clap\_general & 194M & 512 & 35.0 & 41.1 & 18.9 & 47.9 & 6.6 & 80.6 & 51.9 & 66.8 & 14.9 \\
jina-embed-v5-omni-nano & 986M & 768 & \textbf{50.1} & \textbf{55.3} & 50.0 & 51.3 & 6.0 & 70.8 & 65.6 & 83.2 & \textbf{60.3} \\
larger\_clap\_music\_and\_speech & 194M & 512 & 34.7 & 40.7 & 18.6 & 47.4 & 7.7 & 80.6 & 52.1 & 65.6 & 13.2 \\
clap-htsat-unfused & 153M & 512 & 33.0 & 40.5 & 18.1 & 41.9 & 12.5 & 80.8 & 52.6 & 66.5 & 11.3 \\
msclap-2023 & 160M & 1024 & 34.0 & 42.5 & 18.7 & 42.2 & 15.2 & 79.7 & 53.6 & 75.4 & 12.6 \\
clap-htsat-fused & 154M & 512 & 33.5 & 41.3 & 17.6 & 41.2 & \textbf{22.7} & \textbf{81.1} & 52.0 & 61.3 & 13.2 \\
MuQ-MuLan-large & 630M & 512 & 28.2 & 34.0 & 11.7 & 38.8 & 4.3 & 33.1 & 51.9 & \textbf{85.4} & 12.6 \\
wav2clip & 163M & 512 & 26.6 & 31.1 & 9.5 & 37.4 & 6.0 & 31.8 & 53.6 & 68.9 & 10.8 \\
msclap-2022 & 196M & 1024 & 32.2 & 39.3 & 21.7 & 36.1 & 19.9 & 70.9 & 51.7 & 62.9 & 12.1 \\
speecht5\_multimodal & 298M & 768 & 26.4 & 30.0 & 7.9 & 38.5 & 1.1 & 32.5 & 57.9 & 56.5 & 15.9 \\
larger\_clap\_music & 194M & 512 & 20.7 & 22.8 & 9.1 & 26.7 & 5.3 & 8.1 & 50.3 & 48.6 & 11.5 \\
\bottomrule
\end{tabular}
}
\caption{Performance of top-ranking models under 1B parameters on MAEB.}
\label{tab:leaderboard_audio}
\end{table}

MAEB includes 30 tasks encompassing five acoustic domains: speech, music, environmental, bioacoustics, and emotion. These are spread across unimodal and cross-modal tasks within retrieval, reranking, clustering, and multiple types of classification. MSEB(Retrieval) contributes a focus on multilingual speech understanding. It evaluates the model's ability to retrieve a text passage from a spoken query, both in the same language and across languages, with a total of 26 languages and regional locales.

On MAEB, EmbeddingGemma 2 again sets a new frontier for efficiency. Notably, among sub-1B models, it achieves state-of-the-art scores on any-to-any retrieval and audio classification (\Cref{tab:leaderboard_audio}), as well as on about a third of all individual tasks. It is also state-of-the-art among models of all sizes for classification of audio pairs and a sixth of all individual tasks. Its mean task score is only about 0.2 points away from the best model under 1B parameters, and about 3.5 points away from the highest-scoring model overall. Even its 128-dimensional embeddings beat every other sub-1B model.

On average, the model surpasses various audio-text specialists, but there is an interesting specialization divide between them. While EmbeddingGemma 2 does beat specialists like CLAP at non-speech tasks like clustering vehicle sounds (\texttt{VehicleSoundClustering}), these models still outperform in tasks like audio captioning (\texttt{ClothoT2ARetrieval}). In speech tasks, EmbeddingGemma 2 sets the state of the art even in tasks with multiple languages (\texttt{FleursT2ARetrieval}) and singing (\texttt{JamAltLyricA2TRetrieval}). But it sometimes struggles at zero-shot tasks with a paralinguistic focus, despite understanding the acoustics deep down. For example, it struggles tremendously at clustering speakers by gender (\texttt{VoxPopuliGenderClustering}: 0.40), but recovers quickly when doing the same task with a linear probe (\texttt{IEMOCAPGender}: 66.40). This indicates the model's speech representations are dominated by semantics, desirably avoiding biases such as speaker demographic while retaining paralinguistic and acoustic attributes for extraction when needed.

The model's excellence on spoken language is further underscored on MSEB. Even when using ASR and text embeddings, EmbeddingGemma 2 greatly outperforms EmbeddingGemma 1, indicating the strength persists throughout modalities (\Cref{tab:leaderboard_sound}). Encoding audio natively yields another even larger improvement. Instead of suffering from transcription errors from accents, homophones, or artifacts, native audio inputs preserve these signals and let the model take them into account. As a result, even the 256-dimensional embeddings greatly outperform the full-dimension ASR embeddings.
\begin{table}[!t]
\centering
\resizebox{0.85\columnwidth}{!}{%
\begin{tabular}{llccc|cc}
\toprule
Model Name & Audio Mode & Parameters & Dim. & \textbf{Overall} & {In-Lang} & {Cross-Lang} \\
\midrule
{EmbeddingGemma 2} & Native Audio & 577M & 768 & \textcolor{blue}{\textbf{69.54}} & \textcolor{blue}{\textbf{71.63}} & \textcolor{blue}{\textbf{67.68}} \\
\hphantom{EmbeddingGemma 2} & & & 512 & 69.18 & 71.20 & 67.38 \\
\hphantom{EmbeddingGemma 2} & & & 256 & 66.76 & 68.86 & 64.89 \\
\hphantom{EmbeddingGemma 2} & & & 128 & 56.71 & 60.30 & 53.49 \\
\midrule
EmbeddingGemma 2 & ASR & 271M & 768 & 57.59 & 62.14 & 53.52 \\
EmbeddingGemma 1 & ASR & 308M & 768 & 50.95 & 61.10 & 41.87 \\
\midrule\midrule
Gemini Embedding 2 & Native Audio & -- & 3072 & \textbf{73.99} & \textbf{75.58} & \textbf{72.56} \\
Gemini Embedding 2 & ASR & -- & 3072 & 70.40 & 73.58 & 67.55 \\
\bottomrule
\end{tabular}%
}
\caption{Results with ASR and native audio on MSEB(Retrieval) (MRR@10).}
\label{tab:leaderboard_sound}
\vspace{-0.2em}
\end{table}
\vspace{-0.3em}
\section{Future Work}
Our work points to several concrete directions for future research. EmbeddingGemma 2 supports long sequences and variable vision token budgets, but further extending context length, for example via content-aware methods, could enable direct indexing of full-length media, improve performance on long-horizon tasks, and support novel agentic memory systems. Innovations such as encoder-free architectures could enable both quality improvements and further reductions in memory footprint and latency. Changes in training paradigms like the introduction of end-to-end RAG training present another opportunity to further optimize embedding models for downstream use. As interleaved all-to-all multimodal search becomes more prevalent, evaluation frameworks that represent novel use cases will become more important.

At the same time, EmbeddingGemma 2's utility extends far beyond single-vector retrieval, serving as a general-purpose multimodal encoder. Operating on its final embeddings or unpooled token representations, the model can be adapted into (or may even excel zero-shot as) a general decision model \citep{audiojev, llmjev, innerjev, jev}, model router \citep{routers}, reward model \citep{rewardmodels}, reranker \citep{mgte}, speech-to-text model \citep{w2v2, w2vbert}, late interaction retriever \citep{smart, colbert, colpali}, and much more.

\section{Conclusion}
EmbeddingGemma 2 builds on its predecessor with a smaller, stronger backbone and native multimodal capabilities. Our approach introduces improvements like simplified per-layer embeddings and a multi-stage curriculum with both modality and objective progression. This enables EmbeddingGemma 2 to set a new Pareto frontier across text, vision, and audio benchmarks, often surpassing specialist and general-purpose models several times larger. With native support for weight quantization, embedding truncation, and modular encoder loading, EmbeddingGemma 2 makes high-quality multimodal embeddings practical in resource-constrained and on-device settings. We release EmbeddingGemma 2 checkpoints to the community to accelerate research and development, for both off-the-shelf use and as a foundational multimodal encoder.
\newpage
\bibliography{main}

@misc{geminiembedding,
      title={{Gemini Embedding}: Generalizable Embeddings from {Gemini}}, 
      author={Jinhyuk Lee and Feiyang Chen and Sahil Dua and Daniel Cer and Madhuri Shanbhogue and Iftekhar Naim and Gustavo Hernández Ábrego and Zhe Li and Kaifeng Chen and Henrique Schechter Vera and Xiaoqi Ren and Shanfeng Zhang and Daniel Salz and Michael Boratko and Jay Han and Blair Chen and Shuo Huang and Vikram Rao and Paul Suganthan and Feng Han and Andreas Doumanoglou and Nithi Gupta and Fedor Moiseev and Cathy Yip and Aashi Jain and Simon Baumgartner and Shahrokh Shahi and Frank Palma Gomez and Sandeep Mariserla and Min Choi and Parashar Shah and Sonam Goenka and Ke Chen and Ye Xia and Koert Chen and Sai Meher Karthik Duddu and Yichang Chen and Trevor Walker and Wenlei Zhou and Rakesh Ghiya and Zach Gleicher and Karan Gill and Zhe Dong and Mojtaba Seyedhosseini and Yunhsuan Sung and Raphael Hoffmann and Tom Duerig},
      year={2025},
      eprint={2503.07891},
      archivePrefix={arXiv},
      primaryClass={cs.CL},
      url={https://arxiv.org/abs/2503.07891}, 
}

@misc{embeddinggemma,
      title={{EmbeddingGemma: Powerful and Lightweight Text Representations}}, 
      author={Schechter Vera, Henrique and Sahil Dua and Biao Zhang and Daniel Salz and Ryan Mullins and Sindhu Raghuram Panyam and Sara Smoot and Iftekhar Naim and Joe Zou and Feiyang Chen and Daniel Cer and Alice Lisak and Min Choi and Lucas Gonzalez and Omar Sanseviero and Glenn Cameron and Ian Ballantyne and Kat Black and Kaifeng Chen and Weiyi Wang and Zhe Li and Gus Martins and Jinhyuk Lee and Mark Sherwood and Juyeong Ji and Renjie Wu and Jingxiao Zheng and Jyotinder Singh and Abheesht Sharma and Divyashree Sreepathihalli and Aashi Jain and Adham Elarabawy and AJ Co and Andreas Doumanoglou and Babak Samari and Ben Hora and Brian Potetz and Dahun Kim and Enrique Alfonseca and Fedor Moiseev and Feng Han and Frank Palma Gomez and Gustavo Hernández Ábrego and Hesen Zhang and Hui Hui and Jay Han and Karan Gill and Ke Chen and Koert Chen and Madhuri Shanbhogue and Michael Boratko and Paul Suganthan and Sai Meher Karthik Duddu and Sandeep Mariserla and Setareh Ariafar and Shanfeng Zhang and Shijie Zhang and Simon Baumgartner and Sonam Goenka and Steve Qiu and Tanmaya Dabral and Trevor Walker and Vikram Rao and Waleed Khawaja and Wenlei Zhou and Xiaoqi Ren and Ye Xia and Yichang Chen and Yi-Ting Chen and Zhe Dong and Zhongli Ding and Francesco Visin and Gaël Liu and Jiageng Zhang and Kathleen Kenealy and Michelle Casbon and Ravin Kumar and Thomas Mesnard and Zach Gleicher and Cormac Brick and Olivier Lacombe and Adam Roberts and Qin Yin and Yunhsuan Sung and Raphael Hoffmann and Tris Warkentin and Armand Joulin and Tom Duerig and Mojtaba Seyedhosseini},
      year={2025},
      eprint={2509.20354},
      archivePrefix={arXiv},
      primaryClass={cs.CL},
      url={https://arxiv.org/abs/2509.20354}, 
}

@misc{kim2023embeddistillgeometricknowledgedistillation,
      title={{EmbedDistill}: A Geometric Knowledge Distillation for Information Retrieval}, 
      author={Seungyeon Kim and Ankit Singh Rawat and Manzil Zaheer and Sadeep Jayasumana and Veeranjaneyulu Sadhanala and Wittawat Jitkrittum and Aditya Krishna Menon and Rob Fergus and Sanjiv Kumar},
      year={2023},
      eprint={2301.12005},
      archivePrefix={arXiv},
      primaryClass={cs.LG},
      url={https://arxiv.org/abs/2301.12005}, 
}

@inproceedings{kusupati2024matryoshkarepresentationlearning,
 author = {Kusupati, Aditya and Bhatt, Gantavya and Rege, Aniket and Wallingford, Matthew and Sinha, Aditya and Ramanujan, Vivek and Howard-Snyder, William and Chen, Kaifeng and Kakade, Sham and Jain, Prateek and Farhadi, Ali},
 booktitle = {Advances in Neural Information Processing Systems},
 editor = {S. Koyejo and S. Mohamed and A. Agarwal and D. Belgrave and K. Cho and A. Oh},
 pages = {30233--30249},
 publisher = {Curran Associates, Inc.},
 title = {Matryoshka Representation Learning},
 url = {https://proceedings.neurips.cc/paper_files/paper/2022/file/c32319f4868da7613d78af9993100e42-Paper-Conference.pdf},
 volume = {35},
 year = {2022}
}

@INPROCEEDINGS{qat,
  author={Jacob, Benoit and Kligys, Skirmantas and Chen, Bo and Zhu, Menglong and Tang, Matthew and Howard, Andrew and Adam, Hartwig and Kalenichenko, Dmitry},
  booktitle={2018 IEEE/CVF Conference on Computer Vision and Pattern Recognition}, 
  title={Quantization and Training of Neural Networks for Efficient Integer-Arithmetic-Only Inference}, 
  year={2018},
  volume={},
  number={},
  pages={2704-2713},
  doi={10.1109/CVPR.2018.00286}}

@misc{samtone,
      title={{SamToNe}: Improving Contrastive Loss for Dual Encoder Retrieval Models with Same Tower Negatives}, 
      author={Fedor Moiseev and Gustavo Hernandez Abrego and Peter Dornbach and Imed Zitouni and Enrique Alfonseca and Zhe Dong},
      year={2023},
      eprint={2306.02516},
      archivePrefix={arXiv},
      primaryClass={cs.LG},
      url={https://arxiv.org/abs/2306.02516}, 
}

@misc{gemma4,
      title={Gemma 4 Technical Report}, 
      author={Gemma Team},
      year={2026},
      eprint={2607.02770},
      archivePrefix={arXiv},
      primaryClass={cs.CL},
      url={https://arxiv.org/abs/2607.02770}, 
}

@misc{e5v,
      title={{E5-V}: Universal Embeddings with Multimodal Large Language Models}, 
      author={Ting Jiang and Minghui Song and Zihan Zhang and Haizhen Huang and Weiwei Deng and Feng Sun and Qi Zhang and Deqing Wang and Fuzhen Zhuang},
      year={2024},
      eprint={2407.12580},
      archivePrefix={arXiv},
      primaryClass={cs.CL},
      url={https://arxiv.org/abs/2407.12580}, 
}

@misc{vlm2vec,
      title={{VLM2Vec}: Training Vision-Language Models for Massive Multimodal Embedding Tasks}, 
      author={Ziyan Jiang and Rui Meng and Xinyi Yang and Semih Yavuz and Yingbo Zhou and Wenhu Chen},
      year={2025},
      eprint={2410.05160},
      archivePrefix={arXiv},
      primaryClass={cs.CV},
      url={https://arxiv.org/abs/2410.05160}, 
}

@article{spreadout,
  author       = {Xu Zhang and
                  Felix X. Yu and
                  Sanjiv Kumar and
                  Shih{-}Fu Chang},
  title        = {Learning Spread-out Local Feature Descriptors},
  journal      = {CoRR},
  volume       = {abs/1708.06320},
  year         = {2017},
  url          = {http://arxiv.org/abs/1708.06320},
  eprinttype    = {arXiv},
  eprint       = {1708.06320},
  bibsource    = {dblp computer science bibliography, https://dblp.org}
}

@article{same_run_souping,
  author       = {Pavel Izmailov and
                  Dmitrii Podoprikhin and
                  Timur Garipov and
                  Dmitry P. Vetrov and
                  Andrew Gordon Wilson},
  title        = {Averaging Weights Leads to Wider Optima and Better Generalization},
  journal      = {CoRR},
  volume       = {abs/1803.05407},
  year         = {2018},
  url          = {http://arxiv.org/abs/1803.05407},
  eprinttype    = {arXiv},
  eprint       = {1803.05407},
  bibsource    = {dblp computer science bibliography, https://dblp.org}
}

@misc{souping,
      title={Model soups: averaging weights of multiple fine-tuned models improves accuracy without increasing inference time}, 
      author={Mitchell Wortsman and Gabriel Ilharco and Samir Yitzhak Gadre and Rebecca Roelofs and Raphael Gontijo-Lopes and Ari S. Morcos and Hongseok Namkoong and Ali Farhadi and Yair Carmon and Simon Kornblith and Ludwig Schmidt},
      year={2022},
      eprint={2203.05482},
      archivePrefix={arXiv},
      primaryClass={cs.LG},
      url={https://arxiv.org/abs/2203.05482}, 
}

@article{muennighoff2023mtebmassivetextembedding,
  author = {Muennighoff, Niklas and Tazi, Nouamane and Magne, Lo{\"\i}c and Reimers, Nils},
  title = {{MTEB:} Massive Text Embedding Benchmark},
  publisher = {arXiv},
  journal={arXiv preprint arXiv:2210.07316},
  year = {2022},
  url = {https://arxiv.org/abs/2210.07316},
  doi = {10.48550/ARXIV.2210.07316},
}

@article{enevoldsen2025mmteb,
  title={{MMTEB:} Massive Multilingual Text Embedding Benchmark},
  author={Enevoldsen, Kenneth and Chung, Isaac and Kerboua, Imene and Kardos, M{\'a}rton and Mathur, Ashwin and Stap, David and Gala, Jay and Siblini, Wissam and Krzemi{\'n}ski, Dominik and Winata, Genta Indra and others},
  publisher = {arXiv},
  journal={arXiv preprint arXiv:2502.13595},
  year={2025},
  url={https://arxiv.org/abs/2502.13595},
  doi = {10.48550/arXiv.2502.13595},
}

@misc{lee2025nvembedimprovedtechniquestraining,
      title={{NV-Embed:} Improved Techniques for Training {LLMs} as Generalist Embedding Models}, 
      author={Chankyu Lee and Rajarshi Roy and Mengyao Xu and Jonathan Raiman and Mohammad Shoeybi and Bryan Catanzaro and Wei Ping},
      year={2025},
      eprint={2405.17428},
      archivePrefix={arXiv},
      primaryClass={cs.CL},
      url={https://arxiv.org/abs/2405.17428}, 
}

@misc{moreira2025nvretrieverimprovingtextembedding,
      title={{NV-Retriever}: Improving text embedding models with effective hard-negative mining}, 
      author={Gabriel de Souza P. Moreira and Radek Osmulski and Mengyao Xu and Ronay Ak and Benedikt Schifferer and Even Oldridge},
      year={2025},
      eprint={2407.15831},
      archivePrefix={arXiv},
      primaryClass={cs.IR},
      url={https://arxiv.org/abs/2407.15831}, 
}

@misc{santhanam2022colbertv2effectiveefficientretrieval,
      title={{ColBERTv2:} Effective and Efficient Retrieval via Lightweight Late Interaction}, 
      author={Keshav Santhanam and Omar Khattab and Jon Saad-Falcon and Christopher Potts and Matei Zaharia},
      year={2022},
      eprint={2112.01488},
      archivePrefix={arXiv},
      primaryClass={cs.IR},
      url={https://arxiv.org/abs/2112.01488}, 
}

@misc{e5,
      title={Text Embeddings by Weakly-Supervised Contrastive Pre-training}, 
      author={Liang Wang and Nan Yang and Xiaolong Huang and Binxing Jiao and Linjun Yang and Daxin Jiang and Rangan Majumder and Furu Wei},
      year={2024},
      eprint={2212.03533},
      archivePrefix={arXiv},
      primaryClass={cs.CL},
      url={https://arxiv.org/abs/2212.03533}, 
}

@article{clip,
  author         = {Alec Radford and Jong Wook Kim and Chris Hallacy and Aditya Ramesh and Gabriel Goh and Sandhini Agarwal and Girish Sastry and Amanda Askell and Pamela Mishkin and Jack Clark and Gretchen Krueger and Ilya Sutskever},
  title        = {Learning Transferable Visual Models From Natural Language Supervision},
  journal      = {CoRR},
  volume       = {abs/2103.00020},
  year         = {2021},
  url          = {https://doi.org/10.48550/arXiv.2103.00020},
  doi          = {10.48550/ARXIV.2103.00020},
  eprinttype   = {arXiv},
  eprint       = {2103.00020},
  bibsource    = {dblp computer science bibliography, https://dblp.org}
}

@article{align,
  author       = {Chao Jia and
                  Yinfei Yang and
                  Ye Xia and
                  Yi{-}Ting Chen and
                  Zarana Parekh and
                  Hieu Pham and
                  Quoc V. Le and
                  Yun{-}Hsuan Sung and
                  Zhen Li and
                  Tom Duerig},
  title        = {Scaling Up Visual and Vision-Language Representation Learning With
                  Noisy Text Supervision},
  journal      = {CoRR},
  volume       = {abs/2102.05918},
  year         = {2021},
  url          = {https://doi.org/10.48550/arXiv.2102.05918},
  doi          = {10.48550/ARXIV.2102.05918},
  eprinttype   = {arXiv},
  eprint       = {2102.05918},
  bibsource    = {dblp computer science bibliography, https://dblp.org}
}

@misc{siglip,
      title={{SigLIP 2}: Multilingual Vision-Language Encoders with Improved Semantic Understanding, Localization, and Dense Features}, 
      author={Michael Tschannen and Alexey Gritsenko and Xiao Wang and Muhammad Ferjad Naeem and Ibrahim Alabdulmohsin and Nikhil Parthasarathy and Talfan Evans and Lucas Beyer and Ye Xia and Basil Mustafa and Olivier Hénaff and Jeremiah Harmsen and Andreas Steiner and Xiaohua Zhai},
      year={2025},
      eprint={2502.14786},
      archivePrefix={arXiv},
      primaryClass={cs.CV},
      url={https://arxiv.org/abs/2502.14786}, 
}

@misc{mmembed,
      title={{MM-Embed:} Universal Multimodal Retrieval with Multimodal {LLMs}}, 
      author={Sheng-Chieh Lin and Chankyu Lee and Mohammad Shoeybi and Jimmy Lin and Bryan Catanzaro and Wei Ping},
      year={2025},
      eprint={2411.02571},
      archivePrefix={arXiv},
      primaryClass={cs.CL},
      url={https://arxiv.org/abs/2411.02571}, 
}

@misc{gemembed2,
      title={{Gemini Embedding 2}: A Native Multimodal Embedding Model from {Gemini}}, 
      author={Madhuri Shanbhogue and Zhe Li and Shanfeng Zhang and Gustavo Hernández Ábrego and Shih-Cheng Huang and Aashi Jain and Daniel Salz and Sonam Goenka and Chaitra Hegde and Ji Ma and Feiyang Chen and Jiaxing Wu and Tanmaya Dabral and Babak Samari and Kevin Poulet and Daniel Cer and Kaifeng Chen and Paul Suganthan and Hui Hui and Jovan Andonov and Philippe Schlattner and Jay Han and Iftekhar Naim and Wing Lowe and Vladimir Pchelin and Albert Yang and Yi-Ting Chen and Zhongli Ding and Grace Zhang and Georg Heigold and Yichang Chen and Antoine Reveillon and Brendan McCloskey and Wenlei Zhou and Dahun Kim and Rui Meng and Emma Wang and Jack Zheng and Halley Fede and Zhen Yang and Keegan Mosley and Brian Potetz and Sahil Dua and Henrique Schechter Vera and Shen Gao and Hesen Zhang and Andreas Hess and Hengxuan Ying and Alberto Montes and Karan Gill and Min Choi and Sebastian Russo and Anja Hauth and Jinhyuk Lee and Michael Boratko and Megan Barnes and Vikram Rao and Claudiu Musat and Cyril Allauzen and Ehsan Variani and Shankar Kumar and Tom Bagby and Junyi Jiao and Yang Gu and Tengxin Li and Ayush Agrawal and Roberto Santana and Dev Nath and Stephen Karukas and Shuoxuan Han and Lucia Loher and Alice Twu and Nidhi Vyas and Siddharth Bhai and Frank Palma Gomez and Wangyuan Zhang and Chaoren Liu and Jizheng Yang and Steve Qiu and Shijie Zhang and Sujay Kulkarni and Sascha Rothe and Sean Nakamoto and Raphael Hoffmann and Zach Gleicher and Yunhsuan Sung and Qin Yin and Tom Duerig and Mojtaba Seyedhosseini},
      year={2026},
      eprint={2605.27295},
      archivePrefix={arXiv},
      primaryClass={cs.CV},
      url={https://arxiv.org/abs/2605.27295}, 
}

@misc{gme,
      title={{GME}: Improving Universal Multimodal Retrieval by Multimodal {LLMs}}, 
      author={Xin Zhang and Yanzhao Zhang and Wen Xie and Mingxin Li and Ziqi Dai and Dingkun Long and Pengjun Xie and Meishan Zhang and Wenjie Li and Min Zhang},
      year={2025},
      eprint={2412.16855},
      archivePrefix={arXiv},
      primaryClass={cs.CL},
      url={https://arxiv.org/abs/2412.16855}, 
}

@misc{omniembednemotron,
      title={{Omni-Embed-Nemotron}: A Unified Multimodal Retrieval Model for Text, Image, Audio, and Video},
      author={Mengyao Xu and Wenfei Zhou and Yauhen Babakhin and Gabriel Moreira and Ronay Ak and Radek Osmulski and Bo Liu and Even Oldridge and Benedikt Schifferer},
      year={2025},
      eprint={2510.03458},
      archivePrefix={arXiv},
      primaryClass={cs.CL},
      url={https://arxiv.org/abs/2510.03458}, 
}

@misc{lcoembed,
      title={Scaling Language-Centric Omnimodal Representation Learning}, 
      author={Chenghao Xiao and Hou Pong Chan and Hao Zhang and Weiwen Xu and Mahani Aljunied and Yu Rong},
      year={2025},
      eprint={2510.11693},
      archivePrefix={arXiv},
      primaryClass={cs.CL},
      url={https://arxiv.org/abs/2510.11693}, 
}

@misc{e5omni,
      title={e5-omni: Explicit Cross-modal Alignment for Omni-modal Embeddings}, 
      author={Haonan Chen and Sicheng Gao and Radu Timofte and Tetsuya Sakai and Zhicheng Dou},
      year={2026},
      eprint={2601.03666},
      archivePrefix={arXiv},
      primaryClass={cs.CL},
      url={https://arxiv.org/abs/2601.03666}, 
}

@misc{ovis,
      title={{Ovis-Embedding}: Pushing the Frontiers of Universal Omni-Modal Embeddings}, 
      author={Embedding Team},
      year={2026},
      eprint={2609.25165},
      archivePrefix={arXiv},
      primaryClass={cs.AI},
      url={https://arxiv.org/abs/2609.25165}, 
}

@misc{vit,
      title={An Image is Worth 16x16 Words: Transformers for Image Recognition at Scale}, 
      author={Alexey Dosovitskiy and Lucas Beyer and Alexander Kolesnikov and Dirk Weissenborn and Xiaohua Zhai and Thomas Unterthiner and Mostafa Dehghani and Matthias Minderer and Georg Heigold and Sylvain Gelly and Jakob Uszkoreit and Neil Houlsby},
      year={2021},
      eprint={2010.11929},
      archivePrefix={arXiv},
      primaryClass={cs.CV},
      url={https://arxiv.org/abs/2010.11929}, 
}

@misc{usm,
      title={Google {USM}: Scaling Automatic Speech Recognition Beyond 100 Languages}, 
      author={Yu Zhang and Wei Han and James Qin and Yongqiang Wang and Ankur Bapna and Zhehuai Chen and Nanxin Chen and Bo Li and Vera Axelrod and Gary Wang and Zhong Meng and Ke Hu and Andrew Rosenberg and Rohit Prabhavalkar and Daniel S. Park and Parisa Haghani and Jason Riesa and Ginger Perng and Hagen Soltau and Trevor Strohman and Bhuvana Ramabhadran and Tara Sainath and Pedro Moreno and Chung-Cheng Chiu and Johan Schalkwyk and Françoise Beaufays and Yonghui Wu},
      year={2023},
      eprint={2303.01037},
      archivePrefix={arXiv},
      primaryClass={cs.CL},
      url={https://arxiv.org/abs/2303.01037}, 
}

@misc{mieb,
      title={{MIEB}: Massive Image Embedding Benchmark}, 
      author={Chenghao Xiao and Isaac Chung and Imene Kerboua and Jamie Stirling and Xin Zhang and Márton Kardos and Roman Solomatin and Noura Al Moubayed and Kenneth Enevoldsen and Niklas Muennighoff},
      year={2025},
      eprint={2504.10471},
      archivePrefix={arXiv},
      primaryClass={cs.CV},
      url={https://arxiv.org/abs/2504.10471}, 
}

@article{transformer,
  author       = {Ashish Vaswani and
                  Noam Shazeer and
                  Niki Parmar and
                  Jakob Uszkoreit and
                  Llion Jones and
                  Aidan N. Gomez and
                  Lukasz Kaiser and
                  Illia Polosukhin},
  title        = {Attention Is All You Need},
  journal      = {CoRR},
  volume       = {abs/1706.03762},
  year         = {2017},
  url          = {https://doi.org/10.48550/arXiv.1706.03762},
  doi          = {10.48550/ARXIV.1706.03762},
  eprinttype   = {arXiv},
  eprint       = {1706.03762},
  bibsource    = {dblp computer science bibliography, https://dblp.org}
}

@misc{conformer,
      title={Conformer: Convolution-augmented Transformer for Speech Recognition}, 
      author={Anmol Gulati and James Qin and Chung-Cheng Chiu and Niki Parmar and Yu Zhang and Jiahui Yu and Wei Han and Shibo Wang and Zhengdong Zhang and Yonghui Wu and Ruoming Pang},
      year={2020},
      eprint={2005.08100},
      archivePrefix={arXiv},
      primaryClass={eess.AS},
      url={https://arxiv.org/abs/2005.08100}, 
}

@misc{gemma3n,
  author       = {Gemma Team},
  title        = {Gemma 3n},
  howpublished = {\url{https://deepmind.google/models/gemma/gemma-3n/}},
  year         = {2025}
}

@article{gelu,
  author       = {Dan Hendrycks and
                  Kevin Gimpel},
  title        = {Bridging Nonlinearities and Stochastic Regularizers with {Gaussian Error Linear Units}},
  journal      = {CoRR},
  volume       = {abs/1606.08415},
  year         = {2016},
  url          = {https://doi.org/10.48550/arXiv.1606.08415},
  doi          = {10.48550/ARXIV.1606.08415},
  eprinttype   = {arXiv},
  eprint       = {1606.08415},
  bibsource    = {dblp computer science bibliography, https://dblp.org}
}

@inproceedings{Xia2008ListwiseAT,
  title={Listwise approach to learning to rank: theory and algorithm},
  author={Fen Xia and Tie-Yan Liu and Jue Wang and Wensheng Zhang and Hang Li},
  booktitle={International Conference on Machine Learning},
  year={2008},
  url={https://api.semanticscholar.org/CorpusID:207168334},
  doi={10.1145/1390156.1390306}
}

@article{how_contextual_are,
  author       = {Kawin Ethayarajh},
  title        = {How Contextual are Contextualized Word Representations? {Comparing} the Geometry of {BERT, ELMo, and GPT-2} Embeddings},
  journal      = {CoRR},
  volume       = {abs/1909.00512},
  year         = {2019},
  url          = {https://doi.org/10.48550/arXiv.1909.00512},
  doi          = {10.48550/ARXIV.1909.00512},
  eprinttype   = {arXiv},
  eprint       = {1909.00512},
  bibsource    = {dblp computer science bibliography, https://dblp.org}
}

@misc{representation_degeneration,
      title={Representation Degeneration Problem in Training Natural Language Generation Models}, 
      author={Jun Gao and Di He and Xu Tan and Tao Qin and Liwei Wang and Tie-Yan Liu},
      year={2019},
      eprint={1907.12009},
      archivePrefix={arXiv},
      primaryClass={cs.CL},
      url={https://arxiv.org/abs/1907.12009}, 
}

@misc{on_sentence_embeddings,
      title={On the Sentence Embeddings from Pre-trained Language Models}, 
      author={Bohan Li and Hao Zhou and Junxian He and Mingxuan Wang and Yiming Yang and Lei Li},
      year={2020},
      eprint={2011.05864},
      archivePrefix={arXiv},
      primaryClass={cs.CL},
      url={https://arxiv.org/abs/2011.05864}, 
}

@article{uniformity,
  author       = {Tongzhou Wang and
                  Phillip Isola},
  title        = {Understanding Contrastive Representation Learning through Alignment
                  and Uniformity on the Hypersphere},
  journal      = {CoRR},
  volume       = {abs/2005.10242},
  year         = {2020},
  url          = {https://doi.org/10.48550/arXiv.2005.10242},
  doi          = {10.48550/ARXIV.2005.10242},
  eprinttype   = {arXiv},
  eprint       = {2005.10242},
  bibsource    = {dblp computer science bibliography, https://dblp.org}
}

@misc{isotropy_postprocessing,
      title={All-but-the-Top: Simple and Effective Postprocessing for Word Representations}, 
      author={Jiaqi Mu and Suma Bhat and Pramod Viswanath},
      year={2018},
      eprint={1702.01417},
      archivePrefix={arXiv},
      primaryClass={cs.CL},
      url={https://arxiv.org/abs/1702.01417}, 
}

@misc{mindthegap,
      title={Mind the Gap: Understanding the Modality Gap in Multi-modal Contrastive Representation Learning}, 
      author={Weixin Liang and Yuhui Zhang and Yongchan Kwon and Serena Yeung and James Zou},
      year={2022},
      eprint={2203.02053},
      archivePrefix={arXiv},
      primaryClass={cs.CL},
      url={https://arxiv.org/abs/2203.02053}, 
}

@misc{jinav5omni,
      title={jina-embeddings-v5-omni: Geometry-preserving Embeddings via Locked Aligned Towers}, 
      author={Florian Hönicke and Michael Günther and Andreas Koukounas and Mohammad Kalim Akram and Saba Sturua and Han Xiao},
      year={2026},
      eprint={2605.08384},
      archivePrefix={arXiv},
      primaryClass={cs.CL},
      url={https://arxiv.org/abs/2605.08384}, 
}

@misc{prismatic,
      title={Prismatic {VLMs}: Investigating the Design Space of Visually-Conditioned Language Models}, 
      author={Siddharth Karamcheti and Suraj Nair and Ashwin Balakrishna and Percy Liang and Thomas Kollar and Dorsa Sadigh},
      year={2024},
      eprint={2402.07865},
      archivePrefix={arXiv},
      primaryClass={cs.CV},
      url={https://arxiv.org/abs/2402.07865}, 
}

@misc{llava,
      title={Visual Instruction Tuning}, 
      author={Haotian Liu and Chunyuan Li and Qingyang Wu and Yong Jae Lee},
      year={2023},
      eprint={2304.08485},
      archivePrefix={arXiv},
      primaryClass={cs.CV},
      url={https://arxiv.org/abs/2304.08485}, 
}

@misc{palmE,
      title={{PaLM-E}: An Embodied Multimodal Language Model}, 
      author={Danny Driess and Fei Xia and Mehdi S. M. Sajjadi and Corey Lynch and Aakanksha Chowdhery and Brian Ichter and Ayzaan Wahid and Jonathan Tompson and Quan Vuong and Tianhe Yu and Wenlong Huang and Yevgen Chebotar and Pierre Sermanet and Daniel Duckworth and Sergey Levine and Vincent Vanhoucke and Karol Hausman and Marc Toussaint and Klaus Greff and Andy Zeng and Igor Mordatch and Pete Florence},
      year={2023},
      eprint={2303.03378},
      archivePrefix={arXiv},
      primaryClass={cs.LG},
      url={https://arxiv.org/abs/2303.03378}, 
}

@misc{uembed,
      title={{UEmbed}: Unified Sparse and Dense Multimodal Embeddings}, 
      author={Tingyu Song and Mingxin Li and Yanzhao Zhang and Dingkun Long and Pengjun Xie and Zhijie Nie and Yilun Zhao and Shu Wu},
      year={2026},
      eprint={2608.02583},
      archivePrefix={arXiv},
      primaryClass={cs.CV},
      url={https://arxiv.org/abs/2608.02583}, 
}

@misc{cambrian,
      title={Cambrian-1: A Fully Open, Vision-Centric Exploration of Multimodal {LLMs}}, 
      author={Shengbang Tong and Ellis Brown and Penghao Wu and Sanghyun Woo and Manoj Middepogu and Sai Charitha Akula and Jihan Yang and Shusheng Yang and Adithya Iyer and Xichen Pan and Ziteng Wang and Rob Fergus and Yann LeCun and Saining Xie},
      year={2024},
      eprint={2406.16860},
      archivePrefix={arXiv},
      primaryClass={cs.CV},
      url={https://arxiv.org/abs/2406.16860}, 
}

@misc{mm1,
      title={{MM1}: Methods, Analysis \& Insights from Multimodal {LLM} Pre-training}, 
      author={Brandon McKinzie and Zhe Gan and Jean-Philippe Fauconnier and Sam Dodge and Bowen Zhang and Philipp Dufter and Dhruti Shah and Xianzhi Du and Futang Peng and Floris Weers and Anton Belyi and Haotian Zhang and Karanjeet Singh and Doug Kang and Ankur Jain and Hongyu Hè and Max Schwarzer and Tom Gunter and Xiang Kong and Aonan Zhang and Jianyu Wang and Chong Wang and Nan Du and Tao Lei and Sam Wiseman and Guoli Yin and Mark Lee and Zirui Wang and Ruoming Pang and Peter Grasch and Alexander Toshev and Yinfei Yang},
      year={2024},
      eprint={2403.09611},
      archivePrefix={arXiv},
      primaryClass={cs.CV},
      url={https://arxiv.org/abs/2403.09611}, 
}

@misc{blip2,
      title={{BLIP-2}: Bootstrapping Language-Image Pre-training with Frozen Image Encoders and Large Language Models}, 
      author={Junnan Li and Dongxu Li and Silvio Savarese and Steven Hoi},
      year={2023},
      eprint={2301.12597},
      archivePrefix={arXiv},
      primaryClass={cs.CV},
      url={https://arxiv.org/abs/2301.12597}, 
}

@misc{mkp,
      title={Multi-modal Knowledge Preserving Adapter for Embedding Backward Compatibility}, 
      author={Jaeseok Byun and Gukyeong Kwon and Han-Kai Hsu and Meher Gitika Karumuri and Zhikang Zhang and Hao Yang and Davide Modolo},
      year={2026},
      eprint={2609.16875},
      archivePrefix={arXiv},
      primaryClass={cs.CV},
      url={https://arxiv.org/abs/2609.16875}, 
}

@misc{universalgeometry,
      title={Harnessing the Universal Geometry of Embeddings}, 
      author={Rishi Jha and Collin Zhang and Vitaly Shmatikov and John X. Morris},
      year={2026},
      eprint={2505.12540},
      archivePrefix={arXiv},
      primaryClass={cs.LG},
      url={https://arxiv.org/abs/2505.12540}, 
}

@misc{llm2vec,
      title={{LLM2Vec}: Large Language Models Are Secretly Powerful Text Encoders}, 
      author={Parishad BehnamGhader and Vaibhav Adlakha and Marius Mosbach and Dzmitry Bahdanau and Nicolas Chapados and Siva Reddy},
      year={2024},
      eprint={2404.05961},
      archivePrefix={arXiv},
      primaryClass={cs.CL},
      url={https://arxiv.org/abs/2404.05961}, 
}

@misc{wiseft,
      title={Robust fine-tuning of zero-shot models}, 
      author={Mitchell Wortsman and Gabriel Ilharco and Jong Wook Kim and Mike Li and Simon Kornblith and Rebecca Roelofs and Raphael Gontijo-Lopes and Hannaneh Hajishirzi and Ali Farhadi and Hongseok Namkoong and Ludwig Schmidt},
      year={2022},
      eprint={2109.01903},
      archivePrefix={arXiv},
      primaryClass={cs.CV},
      url={https://arxiv.org/abs/2109.01903}, 
}

@misc{apprentice,
      title={Apprentice: Using Knowledge Distillation Techniques To Improve Low-Precision Network Accuracy}, 
      author={Asit Mishra and Debbie Marr},
      year={2017},
      eprint={1711.05852},
      archivePrefix={arXiv},
      primaryClass={cs.LG},
      url={https://arxiv.org/abs/1711.05852}, 
}

@misc{mseb,
      title={Massive Sound Embedding Benchmark {(MSEB)}}, 
      author={Georg Heigold and Ehsan Variani and Tom Bagby and Cyril Allauzen and Ji Ma and Shankar Kumar and Michael Riley},
      year={2026},
      eprint={2602.07143},
      archivePrefix={arXiv},
      primaryClass={cs.SD},
      url={https://arxiv.org/abs/2602.07143}, 
}

@misc{maeb,
      title={{MAEB}: Massive Audio Embedding Benchmark}, 
      author={Adnan El Assadi and Isaac Chung and Chenghao Xiao and Roman Solomatin and Animesh Jha and Rahul Chand and Silky Singh and Kaitlyn Wang and Ali Sartaz Khan and Marc Moussa Nasser and Sufen Fong and Pengfei He and Alan Xiao and Ayush Sunil Munot and Aditya Shrivastava and Artem Gazizov and Niklas Muennighoff and Kenneth Enevoldsen},
      year={2026},
      eprint={2602.16008},
      archivePrefix={arXiv},
      primaryClass={cs.SD},
      url={https://arxiv.org/abs/2602.16008}, 
}

@misc{colbert,
      title={{ColBERT}: Efficient and Effective Passage Search via Contextualized Late Interaction over {BERT}}, 
      author={Omar Khattab and Matei Zaharia},
      year={2020},
      eprint={2004.12832},
      archivePrefix={arXiv},
      primaryClass={cs.IR},
      url={https://arxiv.org/abs/2004.12832}, 
}

@misc{smart,
      title={Your Embedding Model is {SMARTer} Than You Think}, 
      author={Jianrui Zhang and Hyun Jung Lee and Sukanta Ganguly and Tae-Eui Kam and Donghyun Kim and Yong Jae Lee},
      year={2026},
      eprint={2605.24938},
      archivePrefix={arXiv},
      primaryClass={cs.IR},
      url={https://arxiv.org/abs/2605.24938}, 
}

@misc{colpali,
      title={{ColPali}: Efficient Document Retrieval with Vision Language Models}, 
      author={Manuel Faysse and Hugues Sibille and Tony Wu and Bilel Omrani and Gautier Viaud and Céline Hudelot and Pierre Colombo},
      year={2025},
      eprint={2407.01449},
      archivePrefix={arXiv},
      primaryClass={cs.IR},
      url={https://arxiv.org/abs/2407.01449}, 
}

@misc{w2vbert,
      title={{W2v-BERT}: Combining Contrastive Learning and Masked Language Modeling for Self-Supervised Speech Pre-Training}, 
      author={Yu-An Chung and Yu Zhang and Wei Han and Chung-Cheng Chiu and James Qin and Ruoming Pang and Yonghui Wu},
      year={2021},
      eprint={2108.06209},
      archivePrefix={arXiv},
      primaryClass={cs.LG},
      url={https://arxiv.org/abs/2108.06209}, 
}

@misc{w2v2,
      title={wav2vec 2.0: A Framework for Self-Supervised Learning of Speech Representations}, 
      author={Alexei Baevski and Henry Zhou and Abdelrahman Mohamed and Michael Auli},
      year={2020},
      eprint={2006.11477},
      archivePrefix={arXiv},
      primaryClass={cs.CL},
      url={https://arxiv.org/abs/2006.11477}, 
}

@misc{innerjev,
      title={Releasing the Inner {Jev}: A Simple Recipe for Open Decision Models}, 
      author={Jonathan Geuter and Andrew Cheng and David Alvarez-Melis},
      year={2026},
      eprint={2610.17205},
      archivePrefix={arXiv},
      primaryClass={cs.AI},
      url={https://arxiv.org/abs/2610.17205}, 
}

@misc{llmjev,
      title={{LLM-as-Jev}: {LLMs} Are Already {Jev}-Style Decision Models -- When and How to Fine-Tune Them}, 
      author={Yinheng Li and Justin Wagle},
      year={2026},
      eprint={2610.02076},
      archivePrefix={arXiv},
      primaryClass={cs.CL},
      url={https://arxiv.org/abs/2610.02076}, 
}

@misc{jev,
  title={Introducing System One Models and Jev},
  author={Almeida, Diogo},
  year={2026},
  howpublished={\url{https://typesafe.ai/blog/introducing-system-one-models-and-jev}}
}

@misc{audiojev,
      title={{Qwen-Audio-Jev}: A Unified Interface for Audio Decisions}, 
      author={Ying Fang and Zhifu Gao and Qing Gu and Jiajun He and Kangqi Jing and Xiangang Li and Yu Pu and Wupeng Wang},
      year={2026},
      eprint={2610.21614},
      archivePrefix={arXiv},
      primaryClass={cs.SD},
      url={https://arxiv.org/abs/2610.21614}, 
}

@misc{mgte,
      title={{mGTE}: Generalized Long-Context Text Representation and Reranking Models for Multilingual Text Retrieval}, 
      author={Xin Zhang and Yanzhao Zhang and Dingkun Long and Wen Xie and Ziqi Dai and Jialong Tang and Huan Lin and Baosong Yang and Pengjun Xie and Fei Huang and Meishan Zhang and Wenjie Li and Min Zhang},
      year={2024},
      eprint={2407.19669},
      archivePrefix={arXiv},
      primaryClass={cs.CL},
      url={https://arxiv.org/abs/2407.19669}, 
}

@misc{routers,
      title={{RouteLLM}: Learning to Route {LLMs} with Preference Data}, 
      author={Isaac Ong and Amjad Almahairi and Vincent Wu and Wei-Lin Chiang and Tianhao Wu and Joseph E. Gonzalez and M Waleed Kadous and Ion Stoica},
      year={2025},
      eprint={2406.18665},
      archivePrefix={arXiv},
      primaryClass={cs.LG},
      url={https://arxiv.org/abs/2406.18665}, 
}

@misc{rewardmodels,
      title={{RewardBench}: Evaluating Reward Models for Language Modeling}, 
      author={Nathan Lambert and Valentina Pyatkin and Jacob Morrison and LJ Miranda and Bill Yuchen Lin and Khyathi Chandu and Nouha Dziri and Sachin Kumar and Tom Zick and Yejin Choi and Noah A. Smith and Hannaneh Hajishirzi},
      year={2024},
      eprint={2403.13787},
      archivePrefix={arXiv},
      primaryClass={cs.LG},
      url={https://arxiv.org/abs/2403.13787}, 
}

@misc{vlm2vec2,
      title={{VLM2Vec-V2}: Advancing Multimodal Embedding for Videos, Images, and Visual Documents}, 
      author={Rui Meng and Ziyan Jiang and Ye Liu and Mingyi Su and Xinyi Yang and Yuepeng Fu and Can Qin and Zeyuan Chen and Ran Xu and Caiming Xiong and Yingbo Zhou and Wenhu Chen and Semih Yavuz},
      year={2025},
      eprint={2507.04590},
      archivePrefix={arXiv},
      primaryClass={cs.CV},
      url={https://arxiv.org/abs/2507.04590}, 
}

@misc{gritlm,
      title={Generative Representational Instruction Tuning}, 
      author={Niklas Muennighoff and Hongjin Su and Liang Wang and Nan Yang and Furu Wei and Tao Yu and Amanpreet Singh and Douwe Kiela},
      year={2025},
      eprint={2402.09906},
      archivePrefix={arXiv},
      primaryClass={cs.CL},
      url={https://arxiv.org/abs/2402.09906}, 
}

@misc{gte,
      title={Towards General Text Embeddings with Multi-stage Contrastive Learning}, 
      author={Zehan Li and Xin Zhang and Yanzhao Zhang and Dingkun Long and Pengjun Xie and Meishan Zhang},
      year={2023},
      eprint={2308.03281},
      archivePrefix={arXiv},
      primaryClass={cs.CL},
      url={https://arxiv.org/abs/2308.03281}, 
}
\appendix
\crefalias{section}{appendix}
\newpage
\section{Task Instructions}
\label{sec:task_instructions}
\begin{table}[h!]
\centering
\frenchspacing
\resizebox{\columnwidth}{!}{
\begin{tabular}{lll}
\toprule
{Use Case} & {Query Task Instruction} & {Target Task Instruction} \\
\midrule
Web / Document Search & \texttt{task: search result | query: \{text\}} & \texttt{title: \{title\} | text: \{text\}} \\
Question Answering & \texttt{task: question answering | query: \{text\}} & \texttt{title: \{title\} | text: \{text\}} \\
Fact Checking & \texttt{task: fact checking | query: \{text\}} & \texttt{title: \{title\} | text: \{text\}} \\
Code Search & \texttt{task: code retrieval | query: \{text\}} & \texttt{title: \{title\} | text: \{text\}} \\
\midrule
Classification & \texttt{task: classification | query: \{text\}} & \textit{Same as query} \\
Clustering & \texttt{task: clustering | query: \{text\}} & \textit{Same as query} \\
Semantic Similarity & \texttt{task: sentence similarity | query: \{text\}} & \textit{Same as query} \\
\bottomrule
\end{tabular}
}
\caption{Task instruction prefixes used for EmbeddingGemma 2. Asymmetric tasks (top) format queries and targets differently, using \texttt{title: none} when no document title is available. Symmetric tasks (bottom) apply the same prefix to all inputs.}
\label{tab:task_instructions}
\end{table}
\section{Multilingual and English Leaderboards}
\label{sec:appendix_leaderboards}
\begin{table}[h!]
\centering
\resizebox{\columnwidth}{!}{
\begin{tabular}{lcccc|ccccccccc}
\toprule
& & & \multirow{2}{*}{\parbox{1cm}{\centering{\textbf{Mean}\\\textbf{(Task)}}}} & \multirow{2}{*}{\parbox{1cm}{\centering \textbf{Mean}\\\textbf{(Type)}}} & \multirow{2}{*}{\parbox{1.2cm}{\centering Bitext\\Mining}} & & & \multirow{2}{*}{\parbox{1.2cm}{\centering Inst.\\Retrieval}} & \multirow{2}{*}{\parbox{1.2cm}{\centering Multi.\\Class.}} & \multirow{2}{*}{\parbox{1.2cm}{\centering Pair\\Class.}} & & & \\
Model Name & Parameters & Dim. & & & & {Class.} & {Clus.} & & & & {Rerank.} & {Retrieval} & {STS} \\
\midrule
{EmbeddingGemma 2} & 271M & 768 & 61.4 & 53.8 & 65.7 & 63.0 & 50.4 & 2.9 & 25.2 & 78.8 & 62.9 & 60.8 & 74.7 \\
\hphantom{EmbeddingGemma 2} &  & 512 & 61.2 & 53.7 & 65.4 & 62.7 & 51.0 & 2.5 & 25.1 & 78.8 & 62.8 & 60.2 & 74.4 \\
\hphantom{EmbeddingGemma 2} &  & 256 & 60.4 & 53.0 & 63.6 & 61.9 & 51.1 & 2.0 & 24.7 & 78.7 & 62.0 & 58.7 & 74.1 \\
\hphantom{EmbeddingGemma 2} &  & 128 & 57.9 & 50.7 & 59.4 & 59.7 & 49.0 & 0.8 & 23.3 & 78.5 & 59.9 & 52.3 & 73.6 \\
\midrule
harrier-oss-v1-0.6b & 596M & 1024 & \textbf{69.0} & \textbf{59.0} & \textbf{82.8} & \textbf{73.9} & 54.0 & 0.8 & 26.4 & \textbf{82.1} & 63.2 & \textbf{70.8} & 77.1 \\
harrier-oss-v1-270m & 268M & 640 & 66.5 & 56.9 & 81.5 & 70.8 & 52.5 & -0.5 & 24.0 & 80.1 & 61.9 & 66.4 & 75.4 \\
Qwen3-Embedding-0.6B & 596M & 1024 & 64.3 & 56.0 & 72.2 & 66.8 & 52.3 & 5.1 & 24.6 & 80.8 & 61.4 & 64.6 & 76.2 \\
jina-embed-v5-text-nano & 212M & 768 & 65.5 & 57.7 & 67.7 & 69.2 & 52.7 & 0.0 & \textbf{41.3} & 81.9 & 64.6 & 63.3 & \textbf{78.2} \\
multilingual-e5-large-instruct & 560M & 1024 & 63.2 & 55.1 & 80.1 & 64.9 & 50.7 & -0.4 & 23.0 & 80.8 & 62.6 & 57.1 & 76.8 \\
F2LLM-v2-0.6B & 596M & 1024 & 62.7 & 55.0 & 70.3 & 64.1 & \textbf{56.6} & 1.4 & 25.2 & 79.0 & \textbf{65.1} & 59.3 & 74.2 \\
embeddinggemma-300m & 308M & 768 & 61.1 & 54.3 & 64.4 & 60.9 & 51.2 & \textbf{5.6} & 24.7 & 81.4 & 63.2 & 62.5 & 74.7 \\
BidirLM-0.6B-Embedding & 596M & 1024 & 60.4 & 52.9 & 67.6 & 62.5 & 49.5 & 1.2 & 24.1 & 78.4 & 62.7 & 59.1 & 70.7 \\
F2LLM-v2-330M & 334M & 896 & 60.8 & 53.3 & 69.8 & 61.7 & 54.7 & -1.7 & 24.6 & 77.8 & 63.3 & 56.5 & 72.7 \\
jina-embeddings-v3 & 572M & 1024 & 58.4 & 50.7 & 65.3 & 58.8 & 45.6 & -1.3 & 18.4 & 79.3 & 57.1 & 55.8 & 77.1 \\
nomic-embed-text-v2-moe & 475M & 768 & 57.6 & 49.6 & 60.7 & 60.4 & 44.5 & -3.9 & 19.6 & 76.3 & 60.3 & 57.2 & 71.0 \\
\bottomrule
\end{tabular}
}
\caption{Performance of top-ranking models under 1B parameters on MTEB(Multilingual, v2).}
\label{tab:leaderboard_multilingual}
\end{table}
\begin{table}[h!]
\centering
\resizebox{\columnwidth}{!}{
\begin{tabular}{lcccc|ccccccc}
\toprule
& & & \multirow{2}{*}{\parbox{1cm}{\centering{\textbf{Mean}\\\textbf{(Task)}}}} & \multirow{2}{*}{\parbox{1cm}{\centering \textbf{Mean}\\\textbf{(Type)}}} & & & \multirow{2}{*}{\parbox{1.2cm}{\centering Pair\\Class.}} & & & & \\
Model Name & Parameters & Dim. & & & {Class.} & {Clus.} & & {Rerank.} & {Retrieval} & {STS} & {Summ.} \\
\midrule
{EmbeddingGemma 2} & 271M & 768 & 68.5 & 63.9 & 88.8 & 55.4 & 86.4 & 46.3 & 51.7 & 83.2 & 35.1 \\
\hphantom{EmbeddingGemma 2} &  & 512 & 68.4 & 63.9 & 88.7 & 55.3 & 86.4 & 46.3 & 51.6 & 83.3 & 35.6 \\
\hphantom{EmbeddingGemma 2} &  & 256 & 67.8 & 63.3 & 88.5 & 55.5 & 86.1 & 45.9 & 49.4 & 83.1 & 34.8 \\
\hphantom{EmbeddingGemma 2} &  & 128 & 65.7 & 61.5 & 87.6 & 54.3 & 85.3 & 45.7 & 43.7 & 82.5 & 31.4 \\
\midrule
jina-embed-v5-text-nano & 212M & 768 & 71.1 & \textbf{65.2} & \textbf{90.1} & 53.5 & 84.7 & 49.2 & 58.8 & \textbf{88.3} & 31.9 \\
Qwen3-Embedding-0.6B & 596M & 1024 & 70.5 & 64.7 & 84.6 & 54.1 & 84.4 & 48.2 & 61.8 & 86.6 & 33.4 \\
Yuan-embedding-2.0-en & 596M & 1024 & \textbf{71.7} & \textbf{65.2} & 84.4 & 51.7 & 82.0 & \textbf{53.3} & \textbf{70.7} & 84.9 & 29.7 \\
embeddinggemma-300m & 308M & 768 & 69.7 & 65.1 & 87.6 & \textbf{56.6} & 87.3 & 47.4 & 55.7 & 83.6 & \textbf{37.6} \\
Nano-Em1-0.6B-v2.1 & 596M & 1024 & 67.9 & 64.1 & 87.6 & 55.4 & \textbf{89.9} & 49.5 & 46.6 & 85.4 & 34.2 \\
Tarka-Embedding-350M-V1 & 354M & 1024 & 69.3 & 63.3 & 88.4 & 55.7 & 84.0 & 47.8 & 55.1 & 84.6 & 27.4 \\
Giga-Embeddings-instruct & 484M & 1024 & 69.5 & 64.4 & 90.0 & 55.2 & 84.8 & 47.9 & 56.3 & 82.3 & 34.1 \\
GIST-large-Embedding-v0 & 335M & 1024 & 66.3 & 62.0 & 78.9 & 48.8 & 86.7 & 48.8 & 54.5 & 84.4 & 31.5 \\
UltraBERT-Embed-150M & 151M & 768 & 67.2 & 62.8 & 81.7 & 49.7 & 85.3 & 47.5 & 57.2 & 83.0 & 34.9 \\
mxbai-embed-large-v1 & 335M & 1024 & 66.3 & 62.0 & 79.1 & 47.5 & 87.2 & 48.1 & 55.4 & 84.4 & 32.6 \\
UAE-Large-V1 & 335M & 1024 & 66.4 & 61.8 & 79.1 & 47.9 & 87.2 & 48.4 & 55.9 & 84.4 & 30.1 \\
\bottomrule
\end{tabular}
}
\caption{Performance of top-ranking models under 1B parameters on MTEB(English, v2).}
\label{tab:leaderboard_english}
\end{table}
\section{Full Results}
\label{sec:full_results}
\begin{table}[h!]
\centering
\resizebox{0.75\columnwidth}{!}{
\begin{tabular}[t]{lc}
\toprule
Task Name & Performance \\
\midrule
AILAStatutes & 32.44 \\
AfriSentiClassification & 45.67 \\
AlloProfClusteringS2S.v2 & 52.31 \\
AlloprofReranking & 76.92 \\
AmazonCounterfactualClassification & 85.83 \\
ArXivHierarchicalClusteringP2P & 63.42 \\
ArXivHierarchicalClusteringS2S & 60.59 \\
ArguAna & 59.88 \\
ArmenianParaphrasePC & 93.36 \\
BUCC.v2 & 98.50 \\
BelebeleRetrieval & 71.91 \\
BibleNLPBitextMining & 12.27 \\
BigPatentClustering.v2 & 33.31 \\
BiorxivClusteringP2P.v2 & 48.83 \\
BornholmBitextMining & 49.41 \\
BrazilianToxicTweetsClassification & 28.35 \\
BulgarianStoreReviewSentimentClassfication & 68.68 \\
CEDRClassification & 50.38 \\
CLSClusteringP2P.v2 & 39.91 \\
CSFDSKMovieReviewSentimentClassification & 35.08 \\
CTKFactsNLI & 78.55 \\
CataloniaTweetClassification & 51.66 \\
Core17InstructionRetrieval & 4.34 \\
CovidRetrieval & 77.22 \\
CyrillicTurkicLangClassification & 68.49 \\
CzechProductReviewSentimentClassification & 58.35 \\
DBpediaClassification & 91.83 \\
DalajClassification & 50.30 \\
DiaBlaBitextMining & 81.71 \\
EstonianValenceClassification & 38.59 \\
FaroeseSTS & 71.58 \\
FilipinoShopeeReviewsClassification & 37.17 \\
FinParaSTS & 24.57 \\
FinancialPhrasebankClassification & 83.89 \\
FloresBitextMining & 54.89 \\
GermanSTSBenchmark & 83.14 \\
GreekLegalCodeClassification & 35.42 \\
GujaratiNewsClassification & 87.44 \\
HALClusteringS2S.v2 & 28.35 \\
HagridRetrieval & 99.10 \\
IN22GenBitextMining & 73.59 \\
IndicCrosslingualSTS & 42.92 \\
IndicGenBenchFloresBitextMining & 86.74 \\
IndicLangClassification & 85.49 \\
IndonesianIdClickbaitClassification & 64.53 \\
IsiZuluNewsClassification & 24.76 \\
ItaCaseholdClassification & 69.09 \\
JSICK & 84.23 \\
KorHateSpeechMLClassification & 11.58 \\
KorSarcasmClassification & 59.80 \\
KurdishSentimentClassification & 64.04 \\
LEMBPasskeyRetrieval & 54.25 \\
LegalBenchCorporateLobbying & 93.02 \\
MIRACLRetrievalHardNegatives & 66.11 \\
MLQARetrieval & 76.82 \\
MacedonianTweetSentimentClassification & 57.97 \\
MalteseNewsClassification & 30.94 \\
MasakhaNEWSClassification & 75.70 \\
MasakhaNEWSClusteringS2S & 46.02 \\
MassiveIntentClassification & 61.52 \\
MedrxivClusteringP2P.v2 & 43.13 \\
MultiEURLEXMultilabelClassification & 4.78 \\
MultiHateClassification & 57.64 \\
NTREXBitextMining & 73.97 \\
NepaliNewsClassification & 95.46 \\
News21InstructionRetrieval & 5.10 \\
\bottomrule
\end{tabular}%
\hspace{3em}%
\begin{tabular}[t]{lc}
\toprule
Task Name & Performance \\
\midrule
NollySentiBitextMining & 37.21 \\
NordicLangClassification & 65.88 \\
NorwegianCourtsBitextMining & 92.98 \\
NusaParagraphEmotionClassification & 45.05 \\
NusaTranslationBitextMining & 71.37 \\
NusaX-senti & 68.86 \\
NusaXBitextMining & 67.19 \\
OdiaNewsClassification & 72.81 \\
OpusparcusPC & 92.38 \\
PAC & 67.16 \\
PawsXPairClassification & 56.38 \\
PlscClusteringP2P.v2 & 72.46 \\
PoemSentimentClassification & 57.04 \\
PolEmo2.0-OUT & 58.10 \\
PpcPC & 88.46 \\
PunjabiNewsClassification & 82.87 \\
RTE3 & 86.63 \\
Robust04InstructionRetrieval & -0.75 \\
RomaniBibleClustering & 40.46 \\
RuBQReranking & 71.05 \\
SCIDOCS & 18.14 \\
SIB200ClusteringS2S & 26.40 \\
SICK-R & 82.19 \\
STS12 & 77.82 \\
STS13 & 86.48 \\
STS14 & 82.75 \\
STS15 & 88.45 \\
STS17 & 85.79 \\
STS22.v2 & 69.91 \\
STSB & 81.34 \\
STSBenchmark & 87.48 \\
STSES & 81.24 \\
ScalaClassification & 50.12 \\
SemRel24STS & 65.43 \\
SentimentAnalysisHindi & 68.23 \\
SinhalaNewsClassification & 69.54 \\
SiswatiNewsClassification & 55.88 \\
SlovakMovieReviewSentimentClassification & 78.44 \\
SpartQA & 14.56 \\
SprintDuplicateQuestions & 96.50 \\
StackExchangeClustering.v2 & 91.55 \\
StackOverflowQA & 89.08 \\
StatcanDialogueDatasetRetrieval & 45.01 \\
SwahiliNewsClassification & 61.29 \\
SwednClusteringP2P & 42.72 \\
SwissJudgementClassification & 57.68 \\
T2Reranking & 66.64 \\
TERRa & 57.02 \\
TRECCOVID & 79.33 \\
Tatoeba & 54.18 \\
TempReasonL1 & 1.32 \\
ToxicConversationsClassification & 91.03 \\
TswanaNewsClassification & 32.61 \\
TweetTopicSingleClassification & 73.01 \\
TwitterHjerneRetrieval & 68.69 \\
TwitterURLCorpus & 86.83 \\
VoyageMMarcoReranking & 61.78 \\
WebLINXCandidatesReranking & 12.79 \\
WikiCitiesClustering & 89.59 \\
WikiClusteringP2P.v2 & 27.05 \\
WikipediaRerankingMultilingual & 88.38 \\
WikipediaRetrievalMultilingual & 89.47 \\
WinoGrande & 57.93 \\
XNLI & 75.95 \\
indonli & 54.59 \\
\bottomrule
\end{tabular}
}
\caption{Results of EmbeddingGemma 2 on MTEB(Multilingual, v2).}
\label{tab:full_results_mteb_multilingual}
\end{table}
\begin{table}[ht]
\centering
\resizebox{0.71\columnwidth}{!}{
\begin{tabular}{lc}
\toprule
Task Name & Performance \\
\midrule
AmazonCounterfactualClassification & 91.10 \\
ArXivHierarchicalClusteringP2P & 63.66 \\
ArXivHierarchicalClusteringS2S & 60.59 \\
ArguAna & 59.95 \\
AskUbuntuDupQuestions & 60.61 \\
BIOSSES & 85.33 \\
Banking77Classification & 92.27 \\
BiorxivClusteringP2P.v2 & 48.85 \\
CQADupstackGamingRetrieval & 56.90 \\
CQADupstackUnixRetrieval & 42.16 \\
ClimateFEVERHardNegatives & 23.08 \\
FEVERHardNegatives & 81.65 \\
FiQA2018 & 32.47 \\
HotpotQAHardNegatives & 78.25 \\
ImdbClassification & 91.03 \\
MTOPDomainClassification & 98.74 \\
MassiveIntentClassification & 88.41 \\
MassiveScenarioClassification & 91.66 \\
MedrxivClusteringP2P.v2 & 42.86 \\
MedrxivClusteringS2S.v2 & 39.54 \\
MindSmallReranking & 32.05 \\
SCIDOCS & 18.16 \\
SICK-R & 82.20 \\
STS12 & 77.82 \\
STS13 & 86.48 \\
STS14 & 82.75 \\
STS15 & 88.46 \\
STS17 & 91.08 \\
STS22.v2 & 67.42 \\
STSBenchmark & 87.48 \\
SprintDuplicateQuestions & 96.50 \\
StackExchangeClustering.v2 & 91.79 \\
StackExchangeClusteringP2P.v2 & 48.23 \\
SummEvalSummarization.v2 & 35.11 \\
TRECCOVID & 79.33 \\
Touche2020Retrieval.v3 & 44.70 \\
ToxicConversationsClassification & 91.03 \\
TweetSentimentExtractionClassification & 66.44 \\
TwentyNewsgroupsClustering.v2 & 48.05 \\
TwitterSemEval2015 & 75.83 \\
TwitterURLCorpus & 86.83 \\
\bottomrule
\end{tabular}%
\hspace{3em}%
\begin{tabular}{lc}
\toprule
Task Name & Performance \\
\midrule
AppsRetrieval & 89.19 \\
COIRCodeSearchNetRetrieval & 86.89 \\
CodeEditSearchRetrieval & 80.52 \\
CodeFeedbackMT & 85.60 \\
CodeFeedbackST & 82.16 \\
CodeSearchNetCCRetrieval & 93.84 \\
CodeSearchNetRetrieval & 90.85 \\
CodeTransOceanContest & 84.55 \\
CodeTransOceanDL & 35.60 \\
CosQA & 53.34 \\
StackOverflowQA & 89.08 \\
SyntheticText2SQL & 72.47 \\
\bottomrule
\end{tabular}
}
\caption{Results of EmbeddingGemma 2 on MTEB(English, v2) (left) and MTEB(Code) (right).}
\label{tab:full_results_mteb_eng_code}
\end{table}
\begin{table}[ht]
\centering
\resizebox{0.75\columnwidth}{!}{
\begin{tabular}{lc}
\toprule
Task Name & Performance \\
\midrule
BeijingOpera & 81.37 \\
BirdCLEF & 16.00 \\
CREMADPairClassification & 52.50 \\
CREMA\_D & 19.67 \\
CREMA\_DClustering & 0.57 \\
ClothoT2ARetrieval & 10.71 \\
CommonLanguageAgeDetection & 17.59 \\
CommonVoiceMini21T2ARetrieval & 93.26 \\
FSD2019Kaggle & 38.72 \\
FleursT2ARetrieval & 97.32 \\
GTZANAudioReranking & 68.48 \\
GTZANGenre & 58.00 \\
GigaSpeechT2ARetrieval & 82.03 \\
IEMOCAPGender & 66.40 \\
JamAltArtistA2ARetrieval & 55.55 \\
JamAltLyricA2TRetrieval & 64.01 \\
MACST2ARetrieval & 6.36 \\
MInDS14 & 90.30 \\
MridinghamTonic & 32.89 \\
NMSQAPairClassification & 98.12 \\
RavdessZeroshot & 13.96 \\
SIBFLEURS & 54.53 \\
SpeechCommandsZeroshotv0.02 & 96.86 \\
SpokenSQuADT2ARetrieval & 72.33 \\
UrbanSound8KT2ARetrieval & 0.47 \\
VehicleSoundClustering & 11.31 \\
VoxCelebSA & 45.17 \\
VoxPopuliAccentPairClassification & 51.71 \\
VoxPopuliGenderClustering & 0.40 \\
VoxPopuliLanguageID & 99.60 \\
\bottomrule
\end{tabular}%
\hspace{3em}%
\begin{tabular}{l r c c}
\toprule
& & \multicolumn{2}{c}{Performance (MRR@10)} \\
\cmidrule(lr){3-4}
\multicolumn{2}{c}{Language} & In-Lang & Cross-Lang \\
\midrule
Arabic (Egypt) & (ar-EG) & 75.51 & 75.45 \\
Arabic (Gulf) & (ar-x-gulf) & 77.02 & 77.61 \\
Arabic (Levant) & (ar-x-levant) & 78.60 & 78.47 \\
Arabic (Maghreb) & (ar-x-maghrebi) & 73.31 & 72.54 \\
Bengali (Bangladesh) & (bn-BD) & 79.19 & 79.06 \\
Bengali (India) & (bn-IN) & 80.01 & 80.17 \\
English (Australia) & (en-AU) & 74.26 & - \\
English (UK) & (en-GB) & 68.29 & - \\
English (India) & (en-IN) & 72.78 & - \\
English (Philippines) & (en-PH) & 68.94 & - \\
English (US) & (en-US) & 72.30 & - \\
Finnish (Finland) & (fi-FI) & 48.55 & 48.40 \\
Gujarati (India) & (gu-IN) & - & 62.58 \\
Hindi (India) & (hi-IN) & - & 65.27 \\
Indonesian (Indonesia) & (id-ID) & 75.24 & - \\
Japanese (Japan) & (ja-JP) & - & 67.09 \\
Kannada (India) & (kn-IN) & - & 61.26 \\
Korean (South Korea) & (ko-KR) & 62.45 & 73.09 \\
Malayalam (India) & (ml-IN) & - & 61.63 \\
Marathi (India) & (mr-IN) & - & 62.42 \\
Russian (Russia) & (ru-RU) & 65.24 & 80.79 \\
Swahili & (sw) & 65.35 & - \\
Tamil (India) & (ta-IN) & - & 54.11 \\
Telugu (India) & (te-IN) & 80.74 & 69.85 \\
Urdu (India) & (ur-IN) & - & 59.24 \\
Urdu (Pakistan) & (ur-PK) & - & 56.81 \\
\bottomrule
\end{tabular}
}
\caption{Results of EmbeddingGemma 2 on MAEB (left) and MSEB(Retrieval) (right).}
\label{tab:full_results_maeb_mseb}
\end{table}
\begin{table}[ht]
\centering
\resizebox{0.75\columnwidth}{!}{
\begin{tabular}[t]{lc}
\toprule
Task Name & Performance \\
\midrule
AROCocoOrder & 57.00 \\
AROFlickrOrder & 56.12 \\
AROVisualAttribution & 77.97 \\
AROVisualRelation & 56.93 \\
BLINKIT2IMultiChoice & 75.87 \\
CIFAR100ZeroShot & 75.61 \\
CIRRIT2IRetrieval & 22.12 \\
CUB200I2IRetrieval & 78.84 \\
CVBenchCount & 59.77 \\
CVBenchDepth & 52.67 \\
CVBenchDistance & 53.67 \\
CVBenchRelation & 53.39 \\
Country211 & 20.83 \\
Country211ZeroShot & 22.73 \\
DTD & 77.50 \\
EuroSAT & 85.90 \\
FER2013ZeroShot & 61.80 \\
FGVCAircraftZeroShot & 45.21 \\
Fashion200kI2TRetrieval & 15.32 \\
Food101ZeroShot & 81.58 \\
GTSRB & 82.86 \\
HatefulMemesI2TRetrieval & 78.21 \\
ImageCoDe & 14.16 \\
ImageNetDog15Clustering & 91.39 \\
InfoSeekIT2TRetrieval & 13.87 \\
NIGHTSI2IRetrieval & 26.16 \\
\bottomrule
\end{tabular}%
\hspace{3em}%
\begin{tabular}[t]{lc}
\toprule
Task Name & Performance \\
\midrule
OVENIT2TRetrieval & 16.00 \\
OxfordPets & 93.32 \\
OxfordPetsZeroShot & 90.81 \\
PatchCamelyon & 69.34 \\
RESISC45 & 89.76 \\
RP2kI2IRetrieval & 73.74 \\
STS13VisualSTS & 77.31 \\
STS15VisualSTS & 85.67 \\
SUN397 & 75.13 \\
StanfordCarsZeroShot & 91.17 \\
TinyImageNetClustering & 71.26 \\
VQA2IT2TRetrieval & 16.60 \\
VidoreDocVQARetrieval & 39.81 \\
VidoreInfoVQARetrieval & 77.34 \\
VidoreShiftProjectRetrieval & 65.31 \\
VidoreSyntheticDocQAAIRetrieval & 87.76 \\
VidoreTabfquadRetrieval & 91.33 \\
VidoreTatdqaRetrieval & 42.26 \\
VisualNewsI2TRetrieval & 16.94 \\
VisualSTS-b-Multilingual & 75.59 \\
VisualSTS17Multilingual & 77.95 \\
WITT2IRetrieval & 65.31 \\
WebQAT2ITRetrieval & 64.00 \\
Winoground & 12.00 \\
XM3600T2IRetrieval & 65.45 \\
\bottomrule
\end{tabular}
}
\caption{Results of EmbeddingGemma 2 on MIEB(lite).}
\label{tab:full_results_mieb_lite}
\end{table}
\begin{table}[ht]
\centering
\resizebox{0.71\columnwidth}{!}{
\begin{tabular}[t]{lc}
\toprule
Task Name & Performance \\
\midrule
A-OKVQA & 35.20 \\
ActivityNetQA & 52.10 \\
Breakfast & 35.57 \\
CIRR & 40.10 \\
Charades-STA & 25.58 \\
ChartQA & 18.90 \\
Country211 & 18.50 \\
DiDeMo & 52.49 \\
DocVQA & 51.50 \\
EDIS & 80.70 \\
EgoSchema & 56.60 \\
FashionIQ & 27.90 \\
GQA & 61.40 \\
HMDB51 & 55.30 \\
HatefulMemes & 51.40 \\
ImageNet-1K & 65.00 \\
ImageNet-A & 39.80 \\
ImageNet-R & 87.40 \\
InfographicsVQA & 22.70 \\
K700 & 54.70 \\
MMLongBench-doc & 45.08 \\
MMLongBench-page-fixed & 46.34 \\
MSCOCO & 63.20 \\
MSCOCO\_i2t & 72.80 \\
MSCOCO\_t2i & 77.70 \\
MSR-VTT & 50.90 \\
MSVD & 67.46 \\
MVBench & 43.80 \\
MomentSeeker & 47.88 \\
N24News & 38.10 \\
NExTQA & 52.74 \\
NIGHTS & 66.70 \\
OK-VQA & 43.40 \\
OVEN & 68.00 \\
ObjectNet & 65.50 \\
Place365 & 35.30 \\
QVHighlight & 78.21 \\
RefCOCO & 82.90 \\
RefCOCO-Matching & 72.40 \\
\bottomrule
\end{tabular}%
\hspace{3em}%
\begin{tabular}[t]{lc}
\toprule
Task Name & Performance \\
\midrule
SUN397 & 67.50 \\
ScienceQA & 26.90 \\
SmthSmthV2 & 59.80 \\
TextVQA & 74.10 \\
UCF101 & 75.90 \\
VATEX & 48.06 \\
VOC2007 & 75.20 \\
ViDoRe\_arxivqa & 70.56 \\
ViDoRe\_biomedical\_lectures\_v2\_multilingual & 46.99 \\
ViDoRe\_docvqa & 36.60 \\
ViDoRe\_economics\_reports\_v2\_multilingual & 53.81 \\
ViDoRe\_esg\_reports\_human\_labeled\_v2 & 78.22 \\
ViDoRe\_esg\_reports\_v2\_multilingual & 54.25 \\
ViDoRe\_infovqa & 78.79 \\
ViDoRe\_shiftproject & 63.54 \\
ViDoRe\_syntheticDocQA\_artificial\_intelligence & 81.78 \\
ViDoRe\_syntheticDocQA\_energy & 78.67 \\
ViDoRe\_syntheticDocQA\_government\_reports & 81.87 \\
ViDoRe\_syntheticDocQA\_healthcare\_industry & 87.60 \\
ViDoRe\_tabfquad & 91.62 \\
ViDoRe\_tatdqa & 34.75 \\
ViDoSeek-doc & 79.50 \\
ViDoSeek-page-fixed & 82.29 \\
Video-MME & 37.89 \\
VisDial & 76.50 \\
VisRAG\_ArxivQA & 68.14 \\
VisRAG\_ChartQA & 72.33 \\
VisRAG\_InfoVQA & 83.26 \\
VisRAG\_MP-DocVQA & 67.67 \\
VisRAG\_PlotQA & 55.52 \\
VisRAG\_SlideVQA & 89.08 \\
Visual7W & 38.50 \\
Visual7W-Pointing & 91.00 \\
VisualNews\_i2t & 60.00 \\
VisualNews\_t2i & 58.50 \\
VizWiz & 46.40 \\
WebQA & 89.70 \\
Wiki-SS-NQ & 71.40 \\
YouCook2 & 17.08 \\
\bottomrule
\end{tabular}
}
\caption{Results of EmbeddingGemma 2 on MMEB-v2.}
\label{tab:full_results_mmeb_v2}
\end{table}

\twocolumn[  
    \begin{@twocolumnfalse}
        \vspace{-1.2em}
        \section*{Contributions and Acknowledgements}
     \end{@twocolumnfalse}
]
\noindent\textbf{Core Contributors} ($^*$: equal contributions)\\
Henrique Schechter Vera$^*$\\
Sahil Dua$^*$\\
Jovan Andonov\\
Sara Smoot\\
Sebastian Russo\\
Ivan Llanos\\
Max Matveev\\
Jae Hyeon Yoo\\
Kat Black\vspace{1em}\\
\textbf{Contributors}\\
Shih-Cheng Huang\\
Josiah Aklilu\\
Maarten Grootendorst\\
Omar Sanseviero\\
Chandramouli Amarnath\\
Ian Ballantyne\\
Zhe Li\\
Daniel Salz\\
Ibrahim Alabdulmohsin\\
Ilya Philippov\\
Kristen Quan\\
Glenn Cameron\\
Edouard Yvinec\\
Luciano Martins\\
Doug Reid\\
Sindhu Raghuram Panyam\\
Weiyi Wang\\
Hengxuan Ying\\
Madhuri Shanbhogue\\
Tanmaya Dabral\\
Shanfeng Zhang\\
Juyeong Ji\\
Renjie Wu\\
Joe Zou\\
Xu Chen\\
Yunan Zhang\\
Yu-Hui Chen\\
Matthew Soulanille\\
Vinod Mamillapalli\\
Yi-Chun Kuo\\
Wai Hon Law\\
Matt Kreileder\\
Alice Zheng\\
Suril Shah\\
Mogan Shieh\\
Pooja Consul\vspace{1em}\\
\noindent\textbf{Gemini Embedding} \textit{(alphabetical order)}\\
Alberto Montes\\
Andreas Hess\\
Andres Sevilla\\
Babak Samari\\
Brendan McCloskey\\
Brian Potetz\\
Daniel Cer\\
Dilip Kumar Margam\\
Feodor Beloborodov\\
Georgios Kathareios\\
Gustavo Hernández Ábrego\\
Hesen Zhang\\
Ji Ma\\
Jianan Lu\\
Jiaxing Wu\\
Keegan Mosley\\
Paul Suganthan\\
Rui Meng\\
Shen Gao\\
Shengcao Cao\\
Wing Lowe\\
Yi-Ting Chen\\
Yichang Chen\\
Zhongli Ding\vspace{1em} \\

\noindent\textbf{Leadership}\\
Kendra Byrne\\
Cormac Brick\\
Olivier Lacombe\\
Qin Yin\\
Tris Warkentin\\
Armand Joulin\\
Mojtaba Seyedhosseini\\
Tom Duerig\\
\onecolumn
\noindent\textbf{Acknowledgements}\\
Anand Rao, Raushan Turganbay, Anton Vlasjuk, Tom Aarsen, Srinivasan (Cheenu) Venkatachary, Tom\'{a}\v{s} I\v{z}o, Raia Hadsell, Zoubin Ghahramani

\end{document}